\documentclass[conference]{IEEEtran}
\usepackage{fancyhdr}
\usepackage[normalem]{ulem}
\usepackage[hyphens]{url}

\usepackage{graphicx} 

\usepackage[numbers]{natbib}

\usepackage{algorithm}
\usepackage{algorithmic}

\usepackage{xspace}

\usepackage[font=small,labelfont=bf]{caption}
\usepackage[caption=false]{subfig}
\usepackage[hidelinks]{hyperref}
\usepackage{xspace}
\usepackage{amsmath}
\usepackage{mathtools}
\usepackage{xcolor}

\usepackage{amsfonts}
\usepackage{amssymb}
\usepackage{pifont}

\usepackage{color}

\newcommand{\old}[1]{}

\newcommand{\vdnn}[0]{\texttt{vDNN}\xspace}
\newcommand{\cdma}[0]{\texttt{cDMA}\xspace}

\newcommand{\fig}[1]{Figure~\ref{#1}}
\newcommand{\sect}[1]{Section~\ref{#1}}
\newcommand{\tab}[1]{Table~\ref{#1}}

\newcommand{\e}[1]{\times 10^{#1}}

\newcommand{\featureIn}[0]{\texttt{X}\xspace}
\newcommand{\featureOut}[0]{\texttt{Y}\xspace}
\newcommand{\gradientIn}[0]{\texttt{dY}\xspace}
\newcommand{\gradientOut}[0]{\texttt{dX}\xspace}

\newcommand{\weight}[0]{\texttt{W}\xspace}

\newcommand{\layer}[1]{layer$_{(#1)}$\xspace}

\newcommand{\avgDensity}[0]{\texttt{AVG$_{density}$}\xspace}
\newcommand{\zlib}[0]{\texttt{zlib}\xspace}
\newcommand{\rle}[0]{\texttt{RLE}\xspace}
\newcommand{\zvc}[0]{\texttt{ZVC}\xspace}
\newcommand{\nchw}[0]{\texttt{NCHW}\xspace}
\newcommand{\nhwc}[0]{\texttt{NHWC}\xspace}
\newcommand{\chwn}[0]{\texttt{CHWN}\xspace}

\usepackage{amsfonts}
\usepackage{amssymb}
\usepackage{pifont}

\newcommand\blfootnote[1]{%
\begingroup
\renewcommand\thefootnote{}\footnote{#1}%
\addtocounter{footnote}{-1}%
\endgroup
}

\ifCLASSINFOpdf
\else
\fi
\begin{document}
\pagenumbering{arabic}

%
\title{Compressing DMA Engine: Leveraging Activation\\Sparsity for Training Deep Neural Networks}

\author{
\IEEEauthorblockN{
Minsoo Rhu\hspace{2em}Mike O'Connor\hspace{2em}Niladrish Chatterjee\hspace{2em}Jeff Pool\hspace{2em}Stephen W. Keckler}
\IEEEauthorblockA{
NVIDIA\\
Santa Clara, CA 95050\\
\texttt{\{mrhu, moconnor, nchatterjee, jpool, skeckler\}@nvidia.com}\\
}
}

%



\maketitle
\thispagestyle{plain}
\pagestyle{plain}

\begin{abstract} 
Popular deep learning frameworks require users to fine-tune their
memory usage so that the training data of a deep neural network (DNN)
fits within the GPU physical memory\@. Prior work tries to
address this restriction by virtualizing the memory usage of DNNs,
enabling both CPU and GPU memory to be utilized for memory
allocations. Despite its merits, virtualizing memory can incur
significant performance overheads when the time needed to copy data
back and forth from CPU memory is higher than the latency to perform
the computations required for DNN forward and backward propagation.
We introduce a high-performance virtualization strategy based on a
``compressing DMA engine'' (cDMA) that drastically reduces the size of
the data structures that are targeted for CPU-side allocations.  The
cDMA engine offers an average $2.6\times$ (maximum $13.8\times$)
compression ratio by exploiting the sparsity inherent in 
offloaded data, improving the performance of virtualized DNNs by an
average $32\%$ (maximum $61\%$). 
\end{abstract}


%
\IEEEpeerreviewmaketitle

\blfootnote{
A version submitted to \texttt{arXiv.org}
}

\section{Introduction}
\label{sect:intro}

Deep neural networks (DNNs) are now the driving technology for
numerous application domains, such as computer vision~\cite{alexnet},
speech recognition~\cite{graves:2005:fpc}, and natural language
processing~\cite{collobert:2011:nlp_from_scratch}.  To facilitate the
design and study of DNNs, a large number of machine learning (ML)
frameworks~\cite{caffe,torch,theano,tensorflow,neon,mxnet,cntk} have
been developed in recent years.  Most of these frameworks have strong
backend support for GPUs\@. Thanks to their high compute power and
memory bandwidth~\cite{cudnn}, GPUs can train DNNs orders of magnitude
faster than CPUs\@.  One of the key limitations of these frameworks
however is that the limited physical memory capacity of the GPUs
constrains the algorithm (e.g., the DNN layer width and depth) that
can be trained.

To overcome the GPU memory capacity bottleneck of DNN training, prior
work proposed to \emph{virtualize} the memory usage of DNNs (\vdnn)
such that ML researchers can train larger and deeper neural networks
beyond what is afforded by the physical limits of GPU
memory~\cite{rhu:2016:vdnn}.  By copying GPU-side memory allocations
in and out of CPU memory via the PCIe link, \vdnn exposes both CPU and
GPU memory concurrently for memory allocations which improves user
productivity and flexibility in studying DNN algorithms (detailed in
\sect{sect:motivation}).  However, in certain situations, this
memory-scalability comes at the cost of performance overheads
resulting from the movement of data across the PCIe link.  When the
time needed to copy data back and forth through PCIe is smaller than
the time the GPU spends computing the DNN forward and backward
propagation operations, \vdnn does not affect performance. For
networks whose memory copying operation is bottlenecked by the data
transfer bandwidth of PCIe however, \vdnn can incur significant
overheads with an average $31\%$ performance loss (worst case $52\%$,
\sect{sect:motivation}).  The trend in deep learning is to employ
larger and deeper networks that leads to large memory footprints that
oversubscribe GPU
memory~\cite{vggnet,googlenet,nn_stochastic_depth}.
Therefore, ensuring the performance scalability of the virtualization
features offered by \vdnn is vital for the continued success of deep
learning training on GPUs\@.

Our goal is to develop a virtualization solution for DNNs that
simultaneously meets the dual requirements of memory-scalability and
high performance. To this end, we present a \emph{compressing DMA
  engine} (\cdma), a general purpose DMA architecture for GPUs that
alleviates PCIe bottlenecks by reducing the size of the data
structures copied in and out of GPU memory. Our \cdma architecture
minimizes the design overhead by extending the (de)compression units
already employed in GPU memory controllers as follows.  First, \cdma
requests the memory controller to fetch data from the GPU memory at a
high enough rate (i.e., effective PCIe bandwidth $\times$ compression
ratio) so that the compressed data can be generated at a throughput
commensurate with the PCIe bandwidth. The \cdma copy-engine then
initiates an on-the-fly compression operation on that data, streaming
out the final compressed data to the CPU memory over PCIe\@.  The key
insight derived from our analysis in this paper is that the data
(specifically the \emph{activation} maps of each DNN layer) that are
copied across PCIe contain significant \emph{sparsity} (i.e., fraction
of activations being \emph{zero}-valued) and are highly compressible.
Such sparsity of activations primarily comes from the
ReLU~\cite{alexnet} layers that are extensively used in DNNs\@. We
demonstrate sparsity as well as compressibility of the activation maps
through a data-driven application characterization study.  Overall,
this paper provides a detailed analysis of data sparsity in DNN
activations during training and how our compression pipeline can be
used to overcome the data transfer bandwidth bottlenecks of
virtualized DNNs\@. We show that \cdma provides an average $2.6\times$
(max $13.8\times$) compression ratio and improves performance by an
average $32\%$ (max $61\%$) over the previous \vdnn approach.

\section{Background}
\label{sect:background}

\subsection{Deep Neural Networks}
\label{sect:dnn_arch}

Today's most popular deep neural networks can broadly be categorized
as convolutional neural networks (CNNs) for image recognition, or
recurrent neural networks (RNNs) for video captioning, speech
recognition, and natural language processing. Both CNNs and RNNs are
designed using a combination of multiple types of layers, most notably
the convolutional layers (CONV), activation layers (ACTV), pooling
layers (POOL), and fully-connected layers (FC)\@. A deep neural
network is divided into two functional modules: (a) the \emph{feature
  extraction layers} that learn to extract meaningful features out of
an input, and (b) the \emph{classification layers} that use the
extracted features to analyze and classify the input to a
pre-designated output category.  ``Deep learning'' refers to recent
research trends where a neural network is designed using a large
number of feature extraction layers to learn a deep hierarchy of
features.  The feature extraction layers of a CNN are generally
composed of CONV/ACTV/POOL layers whereas the classification layers
are designed using FC/ACTV layers.

{\bf Convolutional layers.} A convolutional layer contains a
set of filters to identify meaningful features in the input data. 
For visual data such as images, $2$-dimensional filters (or $3$-dimensional when
accounting for the multiple input channels within the input image) are employed
which slide over the input of a layer to perform the convolution operation.

{\bf Activation layers.} An activation layer applies an element-wise
activation function (e.g., \texttt{sigmoid}, \texttt{tanh}, and
ReLU~\cite{alexnet}) to the input feature maps.  The ReLU activation function
in particular is known to provide state-of-the-art performance for CNNs, which
allows positive input values to pass through while thresholding all negative input
values to \emph{zero}.

{\bf Pooling layers.} The pooling layers perform a
spatial-downsampling operation on the input data, resulting in an
output volume that is of smaller size. Downsampling is done via
applying an average or max operation over a region of input elements
and reducing it into a single element.

{\bf Fully-connected layers.} The fully-connected layers (or
classifier layers) constitute the final layers of the network.
Popular choices for FC layers are multi-layer perceptrons, although
other types of FC layers are based on multi-nomial logistic
regression.  The key functionality of this layer type is to find the
correlation between the extracted features and the output category.

\subsection{Training versus Inference}
\label{sect:dnn_training}

\begin{figure}[t!] \centering
\includegraphics[width=0.47\textwidth]{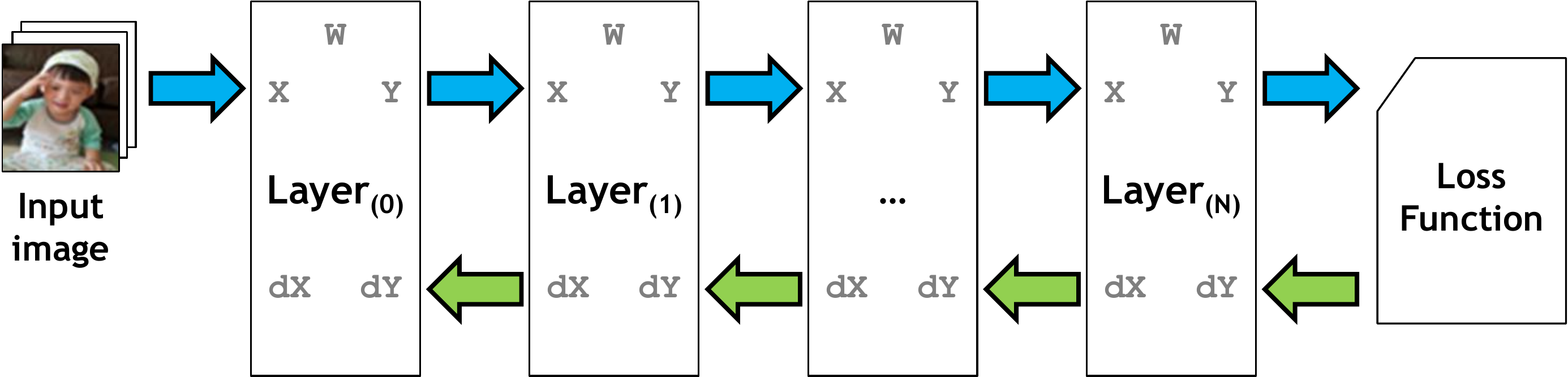}
\caption{
Training a DNN\@.}
\vspace{-1em}
\label{fig:training}
\end{figure}

A neural network requires \emph{training} to be deployed for an
\emph{inference} task. Training a DNN involves learning and updating
the \emph{weights} of the network, which is typically done using the
backpropagation algorithm~\cite{lecun_gd}.  Figure~\ref{fig:training}
shows the three-step process for each training pass: (1) forward
propagation, (2) deriving the magnitude of error between the network's
inference and the ground truth, and (3) propagating the inference
error backwards across the network using backward propagation.

{\bf Forward propagation.} Forward propagation is a serialized,
layer-wise computation process that is performed from the first
(input) layer to the last (output) layer in a sequential manner (from
left to right in \fig{fig:training}).  Each layer applies a
mathematical operation (such as a convolution operation for CONV
layers) to the input \emph{activation maps}\footnote{Following prior
  literature, we refer to the input/output feature maps of any given
  layer as input/output activation maps interchangeably.} (\featureIn)
and generates/stores the results of this operation as output
activation maps (\featureOut).

{\bf Calculating the loss value.} The forward propagation calculation
produces a classification of the input image which must be compared to
the ground truth.  The \emph{loss function} is defined to calculate
the magnitude of this error between classification and ground truth,
deriving the \emph{gradients} of the loss function with respect to the
final layer's output. In general, the loss value drops very quickly at
the beginning of training, and then drops more slowly as the network
becomes fully trained.

{\bf Backward propagation.} Backward propagation is performed in the inverse
direction of forward propagation, from the last layer to the first layer (from
right to left in \fig{fig:training}), again in a layer-wise sequential fashion.
During this phase, the incoming gradients (\gradientIn) can conceptually be
thought of as the inputs to this layer which generate output gradients
(\gradientOut) to be sent to the previous layer. Using these gradients, each
layer adjusts its own layer's weights (\weight), if any (e.g., CONV and FC
layers), so that for the next training pass, the overall loss value is
incrementally reduced. 

With sufficient training examples, which may number in the millions,
the network becomes incrementally better at the task it is trying
to learn.  A detailed discussion of the backpropagation algorithm and
how contemporary GPUs implement each layer's DNN computations and
memory allocations can be found
in~\cite{chetlur:2014:cudnn,rhu:2016:vdnn}.

\subsection{Data Layout for Activation Maps} 
\label{sect:data_layout} 

For training CNNs, the (input/output) activation maps are organized
into a $4$-dimensional array; the number of images batched together
(\texttt{N}), the number of feature map channels per image
(\texttt{C}), and the height (\texttt{H}) and width (\texttt{W}) of
each image. Because the way this $4$-dimensional array is arranged in
memory address space has a significant effect on data locality,
different ML frameworks optimize the layout of their activation maps
differently. For instance, the CNN backend library for
Caffe~\cite{caffe} is optimized for \nchw (i.e., the \texttt{N} and
\texttt{W} in the outermost and innermost dimension of the array,
respectively) whereas cuDNN~\cite{cudnn} provides support for both
\nchw and \nhwc. Neon~\cite{neon} and
\texttt{cuda-convnet}~\cite{cuda_convnet} on the other hand is
optimized for \chwn.  We elaborate on the sensitivity of our proposal
on activation data layout in \sect{sect:compression_eff}.

\section{Motivation}
\label{sect:motivation}

\begin{figure}[t!] \centering
\subfloat[]{
\includegraphics[width=0.25\textwidth]{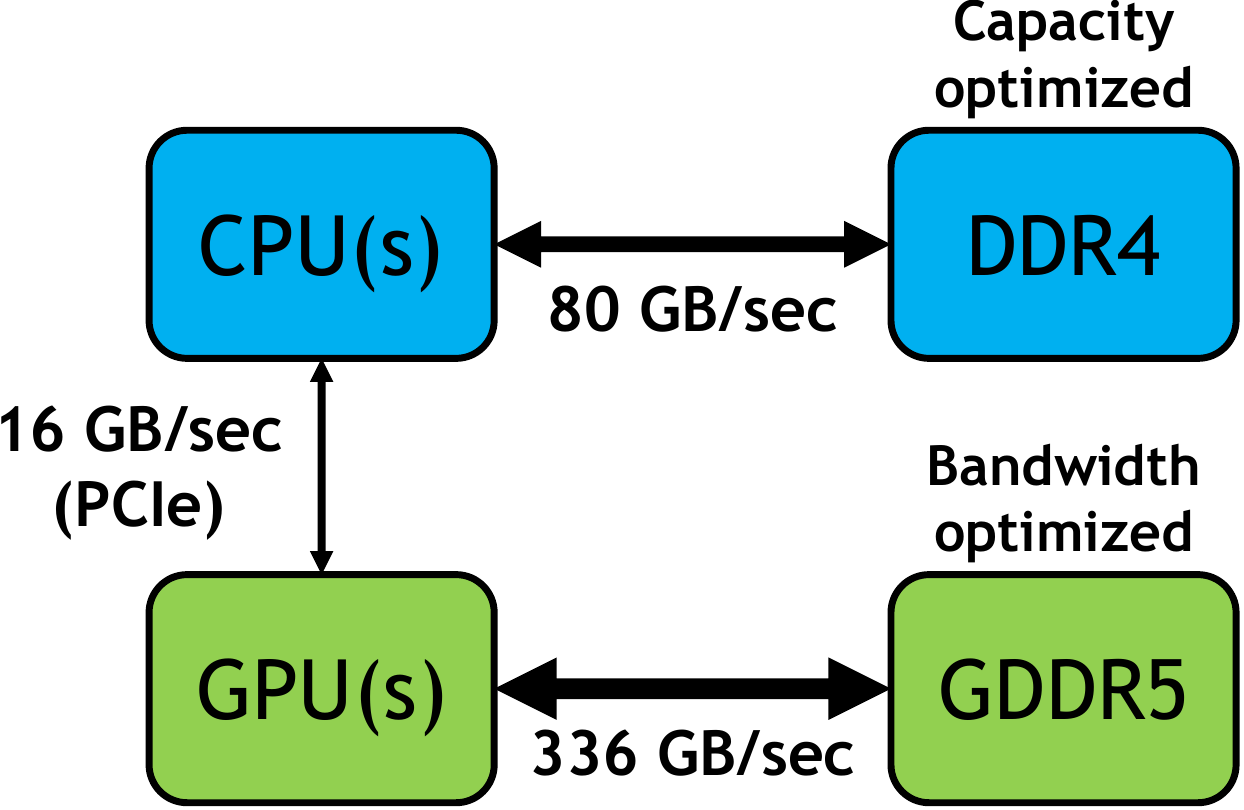}
\label{fig:vdnn_arch}
}
\vspace{0em}
\subfloat[]{
\includegraphics[width=0.49\textwidth]{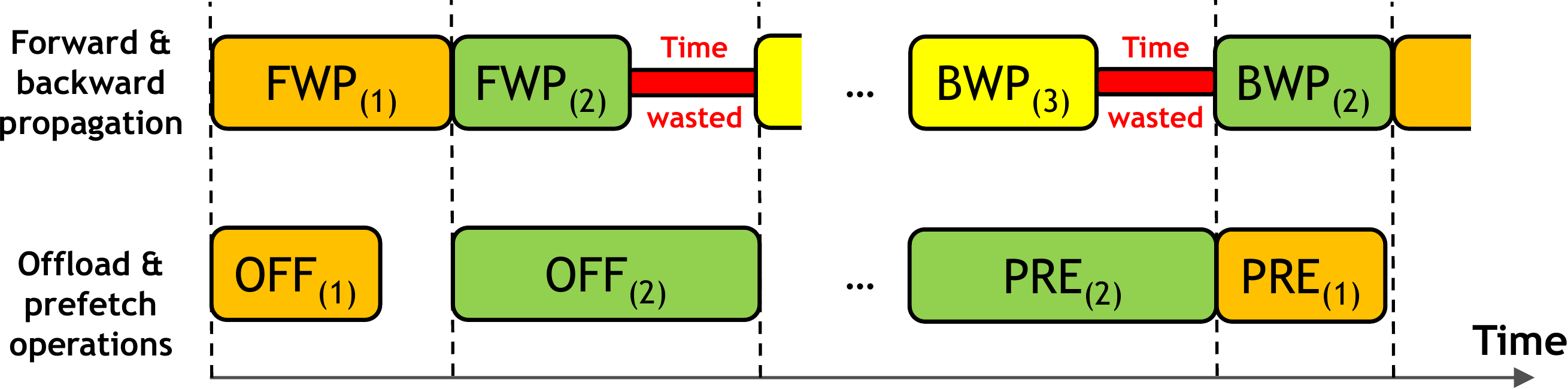}
\label{fig:vdnn_timeline}
}
\caption{(a) PCIe attached CPUs and GPUs, and (b) \vdnn memory management using (GPU-to-CPU) offload and (CPU-to-GPU) prefetch operations.
OFF$_{(n)}$ and PRE$_{(n)}$ corresponds to the offload and prefetch operations of \layer{n}'s activation maps, respectively. } 
\vspace{-1em}
\label{fig:vdnn}
\end{figure}

Several techniques have been proposed for supporting virtual memory on
GPUs\@.  Pichai et al.~\cite{gpu_tlb} and Power et al.~\cite{gpu_x86_at}
proposed TLB designs that leverage the unique memory access patterns
of GPUs for optimizing the throughput of memory address
translations. Zheng et al.~\cite{gpu_paging} studied architectural
solutions for closing the performance gap between page-migration based
virtual memory and software-directed direct-memory-access (DMA) copy
operations. Nonetheless, the performance overheads of these
fine-grained, page-based virtual memory solutions are high because of
the low throughput and high latency of page-migration on discrete GPU
systems. Rhu et al.~\cite{rhu:2016:vdnn} therefore proposed an
application-level virtual memory management solution specifically
tailored for DNNs (\vdnn)\@.  \fig{fig:vdnn} provides a high-level
overview of a state-of-the-art DNN training platform containing CPUs
and GPUs, and how the \vdnn memory manager orchestrates the data copy
operations across the CPU and GPU memory. 
\fig{fig:vdnn}(a) illustrates a discrete GPU card (Maxwell Titan-X)
with $336$ GB/sec of GPU DRAM bandwidth, connected to a host CPU via a
PCIe channel, which provides a maximum data transfer bandwidth of $16$
GB/sec for PCIe gen3.

\fig{fig:vdnn_timeline} shows how \vdnn virtualizes memory by
proactively offloading the inter-layer \emph{activation maps} out to
CPU memory during forward propagation and later prefetching them back
into the GPU, just before they are reused during backward
propagation. For training DNNs, these activation maps occupy more than
$90$\% of the GPU-side memory allocations~\cite{rhu:2016:vdnn}. Thus
\vdnn offers significant reduction in the average GPU memory usage by
offloading activations to the CPU\@.  \vdnn also provides 
much higher PCIe bandwidth utilization and performance than
page-migration based virtual memory (i.e.,
$12.8$GB/sec~\cite{rhu:2016:vdnn} versus $200$
MB/sec~\cite{gpu_paging}) as the data movements are orchestrated by
GPU's DMA copy-engine. However, when the time needed to move data in
and out of the CPU memory takes longer than the time spent computing
DNN's backpropagation algorithm, \vdnn can incur noticeable
performance overheads by stalling the normal DNN computations.

\begin{figure}[t!] \centering
\subfloat[]{
\includegraphics[width=0.49\textwidth]{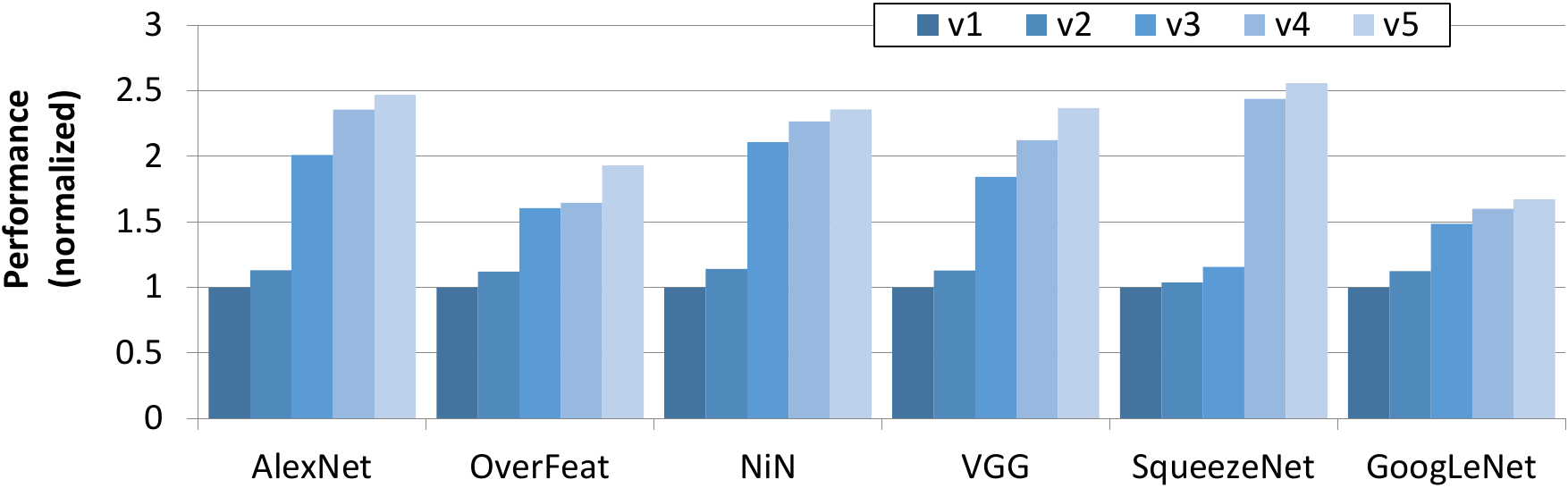}
\label{fig:cudnn_perf_scaling}
}
\vspace{0em}
\subfloat[]{
\includegraphics[width=0.49\textwidth]{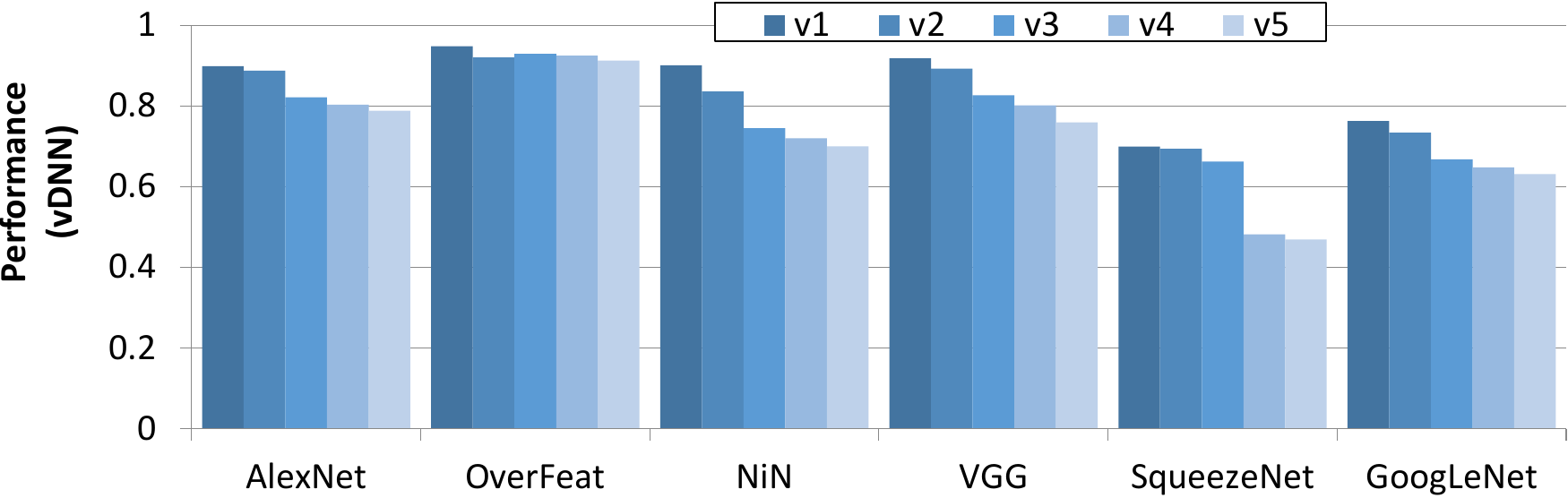}
\label{fig:vdnn_perf}
}
\caption{(a) Speedups offered by versions of cuDNN, and (b) the
performance degradations incurred by {\vdnn}\@.}
\vspace{-1em}
\label{fig:perf_scaling}
\end{figure}

\fig{fig:perf_scaling} illustrates the extent of this bottleneck on
the performance of DNNs\@.  \fig{fig:perf_scaling}(a) shows the
performance improvements offered by successive versions of NVIDIA's
deep learning library cuDNN, which effectively reduces the time spent
computing each CONV layer~\cite{cudnn}.
\fig{fig:perf_scaling}(b) shows the performance overheads imposed by
\vdnn on each of these versions, as the window to overlap the data
transfer with computation diminishes. The most recent version of cuDNN
(v5) offers an average $2.2\times$ the performance of the first
version (v1) released in 2014 across a range of different DNNs\@.
However, the data transfer bandwidth offered by the state-of-the-art
PCIe link (gen3) has remained unchanged at $16$ GB/sec\@. This
divergence is the key reason behind the steadily increasing
performance overheads of \vdnn on successive generations of faster GPU
backend libraries.

Our compressing DMA engine is based on the key observation that the
activation maps, which account for the majority of GPU-side memory
allocations for training deep networks~\cite{rhu:2016:vdnn}, are
amenable for compression, which will drastically alleviate the PCIe
bottleneck of virtualized DNNs\@. A significant fraction of each
layer's activations turn out to be \emph{zero}-valued, meaning these
data structures are \emph{sparse} and are highly compressible.  As
noted by multiple prior
works~\cite{xu:2015:eval_relu,sun:2015:face_sparse,cnvlutin}, such
sparsity of activations are originated by the extensive use of
ReLU~\cite{alexnet} layers that follow (almost) every single layer in
the feature extraction modules. We first provide a data-driven,
in-depth DNN characterization study in \sect{sect:characterization}
that motivates our work, followed by our compressing DMA architecture
in \sect{sect:dma}.  As the effectiveness of our proposition (i.e.,
compression) is highly correlated with the actual data that are input
into the neural network, this paper primarily focuses on convolutional
neural networks (CNNs) owing to their publicly available, realistic
datasets (e.g., ImageNet~\cite{imagenet}) for computer vision tasks.
Nonetheless, we believe our proposal is equally applicable for some
popular recurrent neural networks that extensively employ
sparsity-inducing ReLU layers, including the GEMV-based
(general-matrix-vector-multiplication) RNNs employed by Baidu for
speech recognition~\cite{deepspeech_1,deepspeech_2} and language
translation~\cite{persistent_rnn} services. At present, we cannot
study these RNN applications applications as there are no publicly
available training datasets.  \cdma is less well-suited for RNNs based
on LSTMs~\cite{lstm} or GRUs~\cite{gru}, as they employ
\texttt{sigmoid} and \texttt{tanh} activation functions rather than
ReLUs.

\section{Sparsity of DNN Activations}
\label{sect:characterization}

\begin{figure}[t!] \centering
\includegraphics[width=0.49\textwidth]{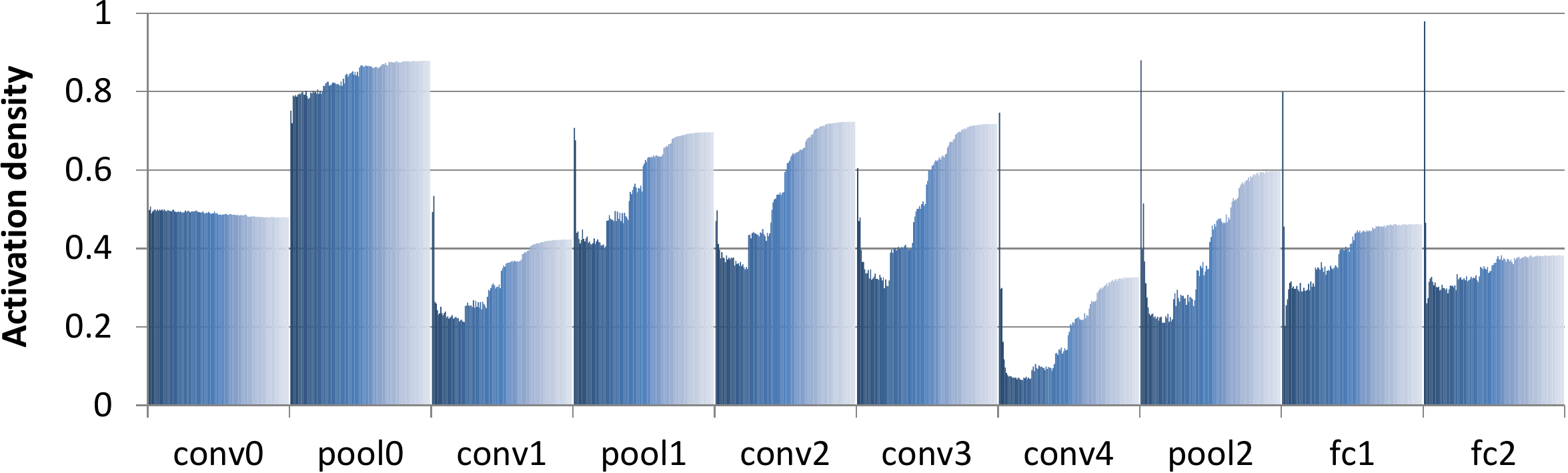}
\vspace*{0em}
\caption{Average activation density of each layer in AlexNet over time during
training (going from dark to light blue colored bars, per layer). Activation
density is sampled at every 2K iterations of training and a total of 226K
iterations were spent to reach the fully trained model (53.1\%/75.1\% top-1/top-5
accuracy).} 
\vspace{-1em}
\label{fig:layer_density_alexnet} 
\end{figure}

The focus of this paper is on DNN training, which involves learning and updating
the weights of a neural network using the backpropagation algorithm.  As
discussed in \sect{sect:dnn_training}, the values in the output activation maps
(\featureOut) are derived as a function of both the input activation maps
(\featureIn) and the layer weights (\weight)\@. The sparsity of each layer's
output activations will therefore change as the training progresses, during
which not only will the layer be continuously fed with new input activation
maps, but the weights for the same layer will undergo changes as a result of
backpropagation. For our compressing DMA engine to be effective, it is crucial that
the activation sparsity, and accordingly its compressibility, remains
consistently high throughout the entire training process. This section analyzes
the effect of training on activation sparsity by using AlexNet~\cite{alexnet} as
a running example. We detail our training methodology in \sect{sect:eval}.

\begin{figure*}[t!] \centering
\includegraphics[width=0.75\textwidth]{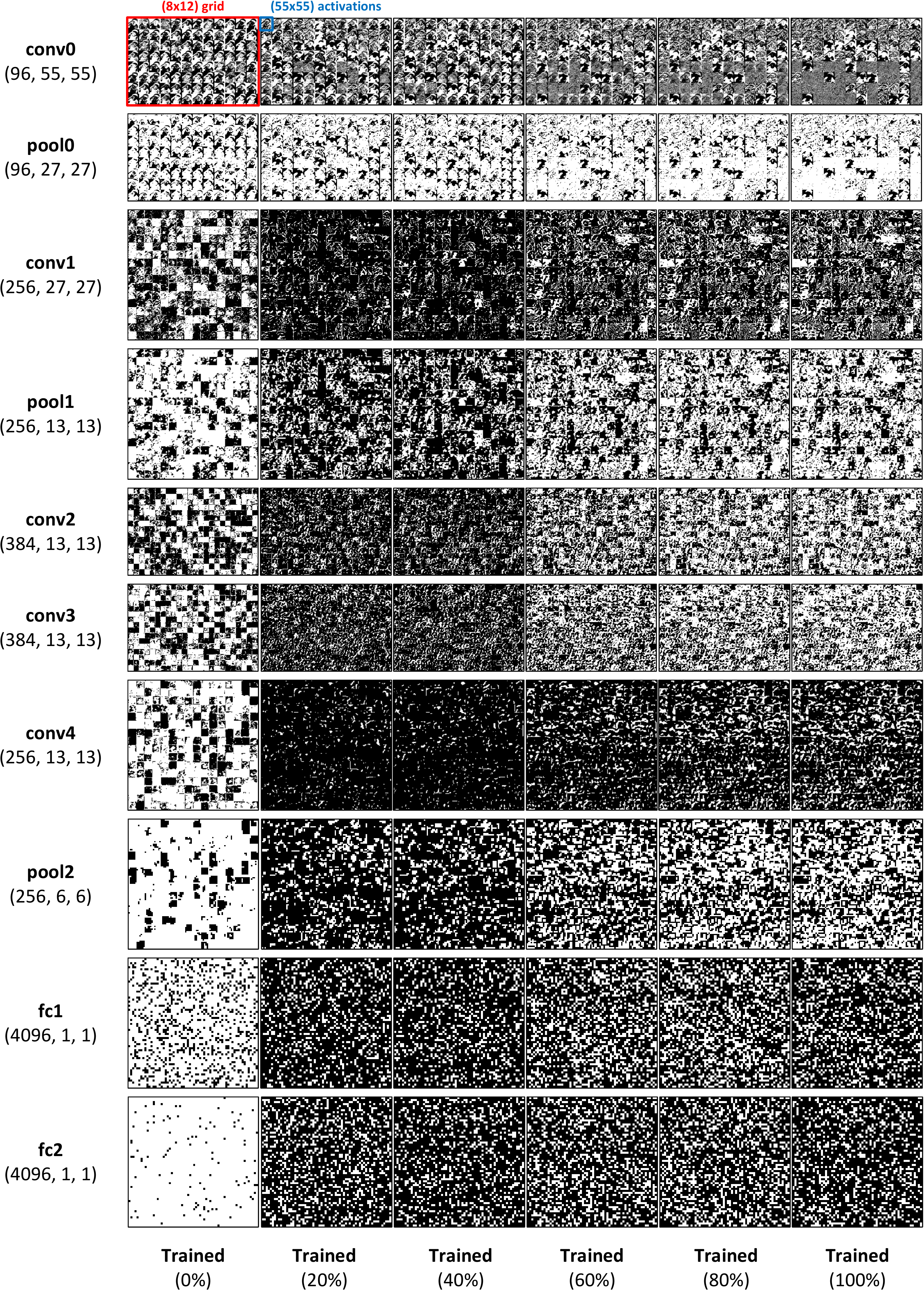} 

\caption{Change in AlexNet's activation sparsity as the network is
  trained.  ``Trained ($0\%$)'' corresponds to point in time when the
  weights of AlexNet were initialized at the onset of training,
  whereas ``Trained ($100\%$)'' designates a fully-trained AlexNet\@.
  The three numbers \texttt{(C, H, W)} below the names of each layer
  represent the number of channels, the height, and the width of the
  output activation maps. A zero value in an activation map is
  represented as a black pixel and white otherwise. The \texttt{RGB}
  image of a boy shown in \fig{fig:training} was used to derive these
  results.
} 
\vspace{-1em}
\label{fig:layer_density_maps} 
\end{figure*}

\subsection{Case Study: Activation Sparsity in AlexNet}
\label{sect:sparsity_alexnet}

\fig{fig:layer_density_alexnet} shows the change in each layer's
average output activation density over time, as the network is trained
for better image classification. We define the per-layer average
output activation density (\avgDensity) as the number of non-zero
output activations divided by the total number of output activations,
which is measured across the minibatch of the same $50$ images.
Accordingly, average activation sparsity is equal to
($1-$\avgDensity).  \fig{fig:layer_density_maps} shows a visualization
of sparsity across time (x-axis), layer (y-axis), and spatially within
each activation map.  For brevity, we only show the layers that are
immediately followed by ReLU layers and would exhibit
sparsity. For instance, the output activation maps of the first
convolutional layer of AlexNet (\texttt{conv0}) contain $96$ channels,
each of which can conceptually be thought of as a $2$-dimensional,
($55\times55$) array of activations per channel.  The $96$ channels
are arranged as a ($8\times12$) grid, with each grid corresponding to
a single channel with ($55\times55$) activations (i.e., the
top-leftmost image in \fig{fig:layer_density_maps}). Each of the
activations are displayed as black and white pixels depending on
whether they are zero-valued (\emph{sparse}, black) or not
(\emph{dense}, white). 

Based on this analysis, we can draw the following key
observations. First, the first convolutional layer (\texttt{conv0}),
regardless of the iterations of training it has gone through, is
neither sparse nor dense, always falling within $\pm2$\% of $50\%$
average activation sparsity (or density). Second, pooling layers
always increase activation density, i.e., activation maps always get
brighter after going through the pooling layers. This result is
expected as pooling layers either pick the highest value (when we use
max pooling) or derive the average value (average pooling) within the
pooling window. Thus, a pooling layer is likely to generate a dense
output unless all the input activations within the pooling window are
all zero-valued. Third, with the exception of the first convolutional
layer, the change in average activation density exhibits a
\texttt{U}-shaped curve during training; the number of non-zero
activations rapidly decreases during the initial training periods but
gradually increases back during the latter stages of training as the
model accuracy improves. This \texttt{U}-shaped curve is also
reflected in \fig{fig:layer_density_maps} where the activation maps
quickly turn extremely dark during the first $40\%$ of the training
period but gradually becoming lighter as the layer enters the
mid-to-end stages of the training process. Finally, layers located
towards the end of the network are generally more sparse than the
earlier layers with the fully-connected layers generally exhibiting
much higher sparsity than the convolutional layers.

Overall, AlexNet exhibits an average $49.4\%$ activation sparsity across the
entire network when accounting for the size of each of the layer's activation
maps (e.g., the sizes of the activation maps in the earlier layers are generally
larger than those located at later layers). Thus, a compression
algorithm that can perfectly compress out all the zeros can reduce the
activation size by about half.

\begin{figure}[t!] \centering
\subfloat[OverFeat]{
\includegraphics[width=0.49\textwidth]{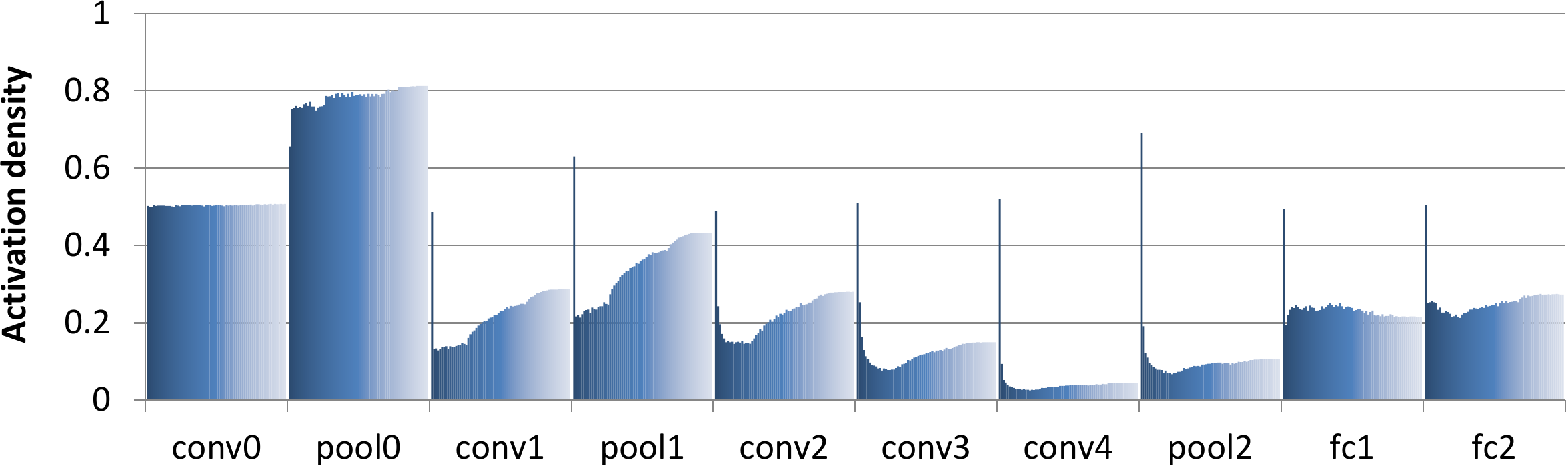}
\label{fig:layer_density_in_time_overfeat}
}
\vspace{0em}
\subfloat[NiN]{
\includegraphics[width=0.49\textwidth]{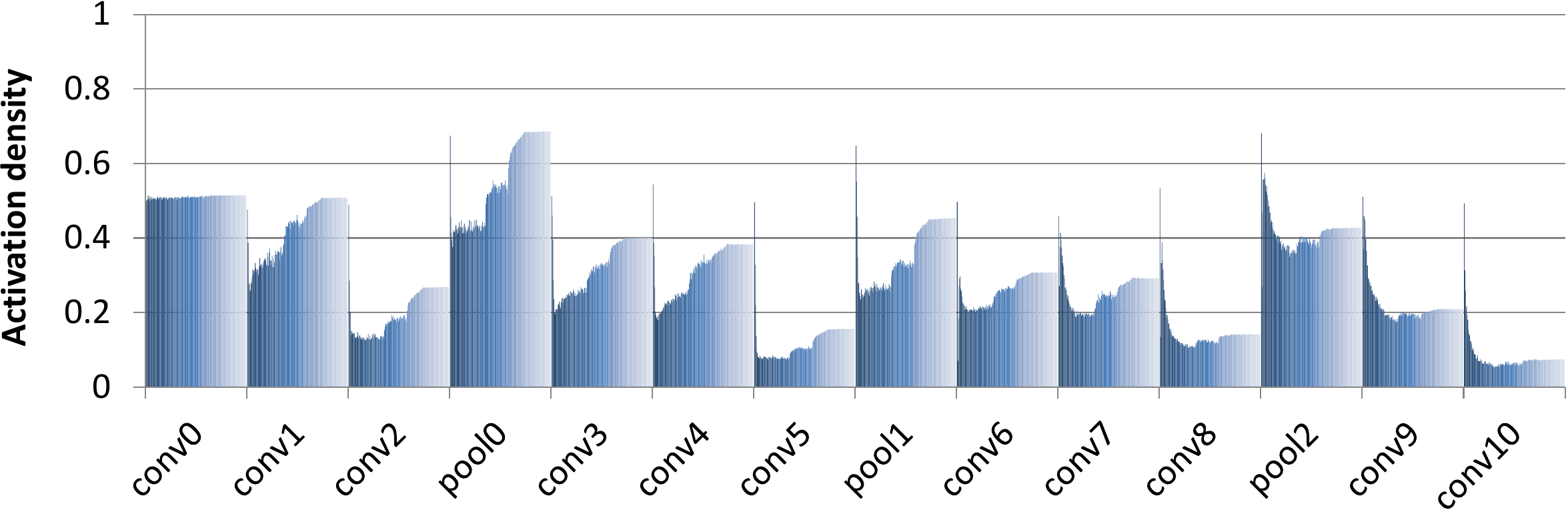}
\label{fig:layer_density_in_time_nin}
}
\vspace{0em}
\subfloat[VGG]{
\includegraphics[width=0.49\textwidth]{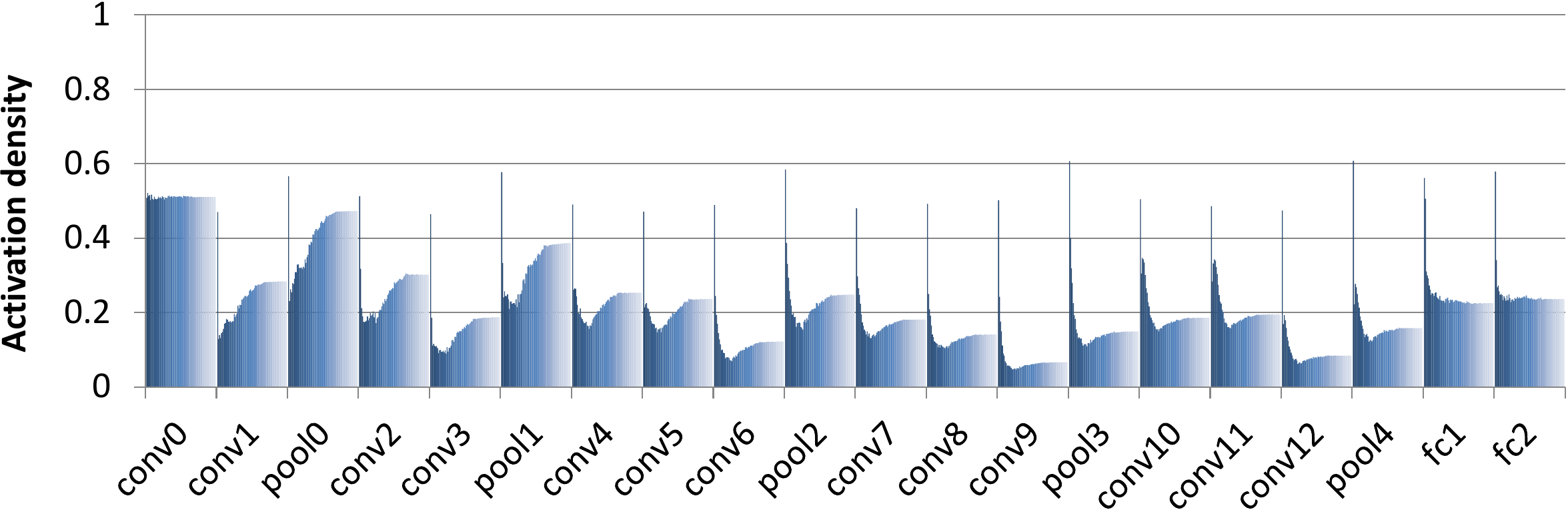}
\label{fig:layer_density_in_time_vgg_16}
}
\vspace{0em}
\subfloat[SqueezeNet]{
\includegraphics[width=0.49\textwidth]{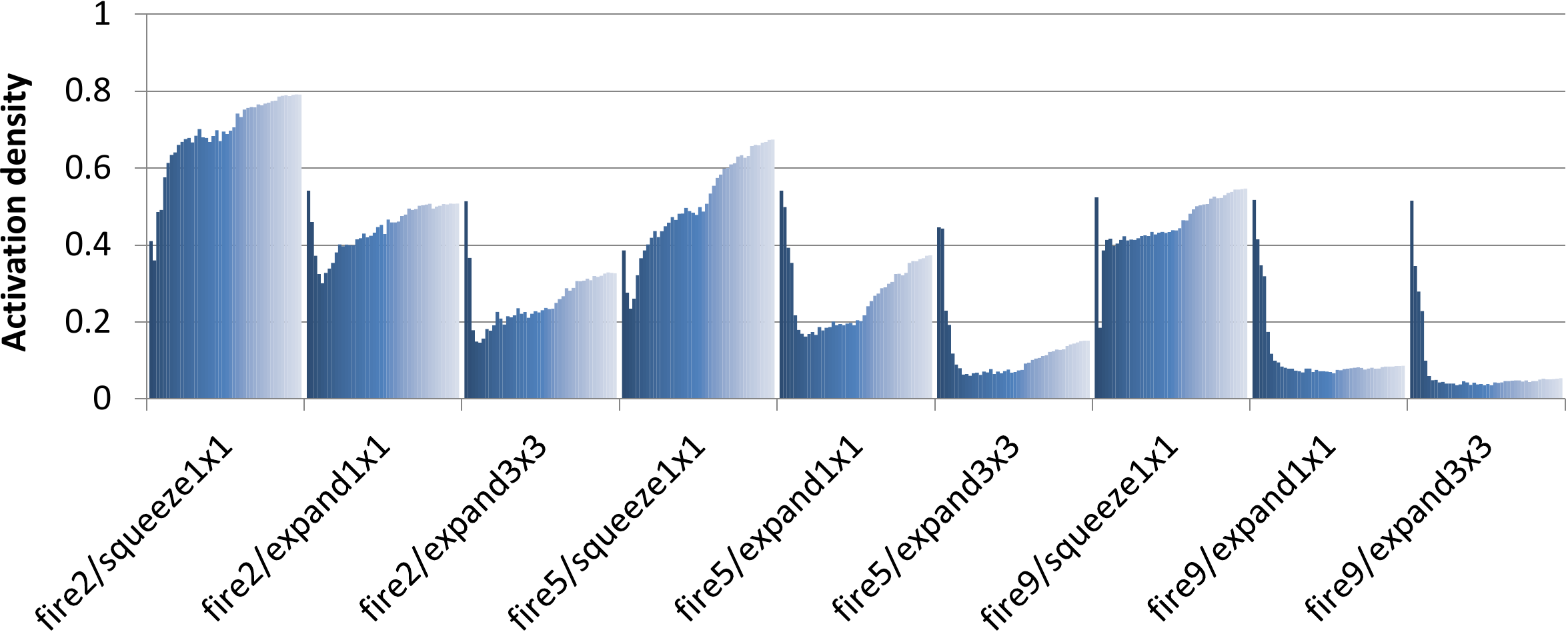}
\label{fig:layer_density_in_time_squeezenet}
}
\vspace{0em}
\subfloat[GoogLeNet]{
\includegraphics[width=0.49\textwidth]{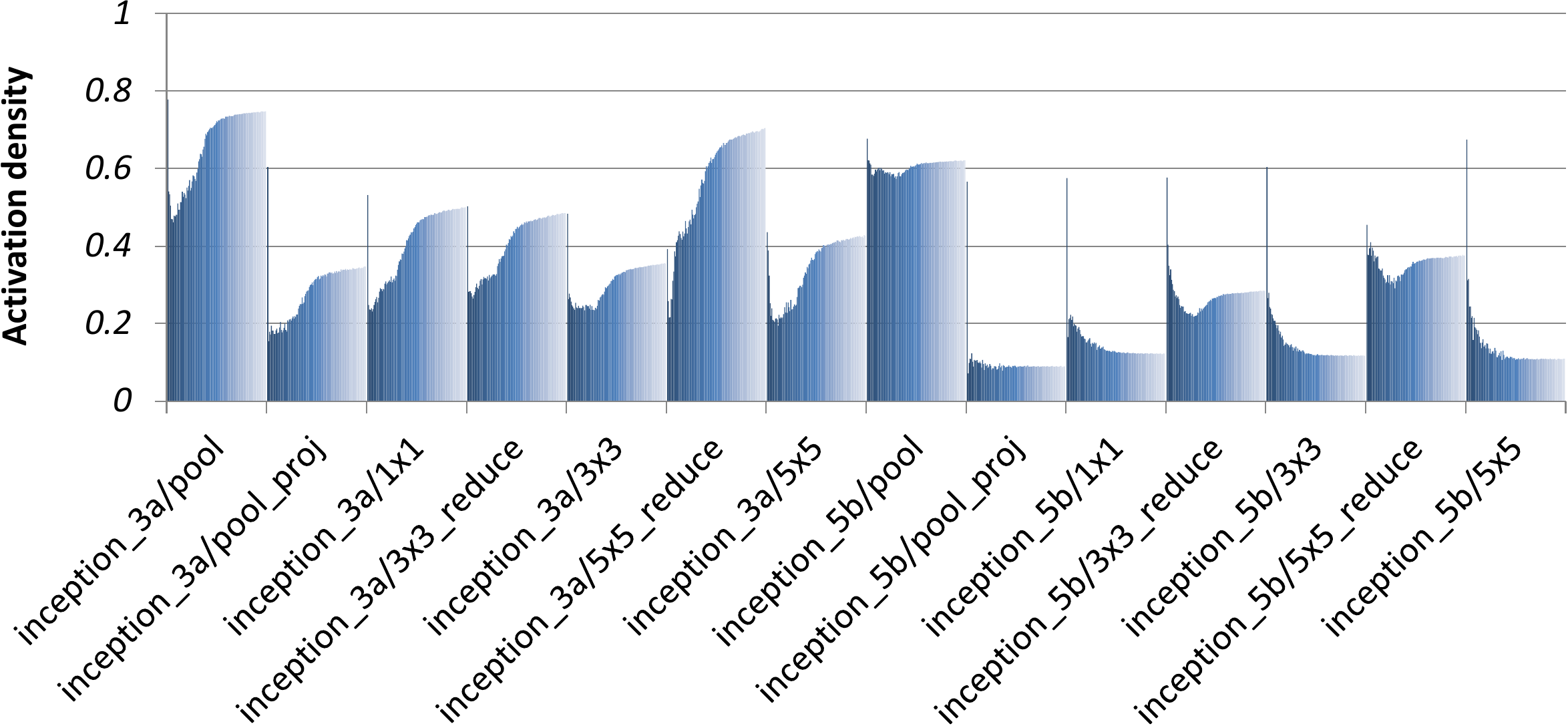}
\label{fig:layer_density_in_time_googlenet}
}
\caption{Effect of training on layer activation density.}
\vspace{-1em}
\label{fig:layer_density} 
\end{figure}

\begin{figure}[t!] \centering
\includegraphics[width=0.49\textwidth]{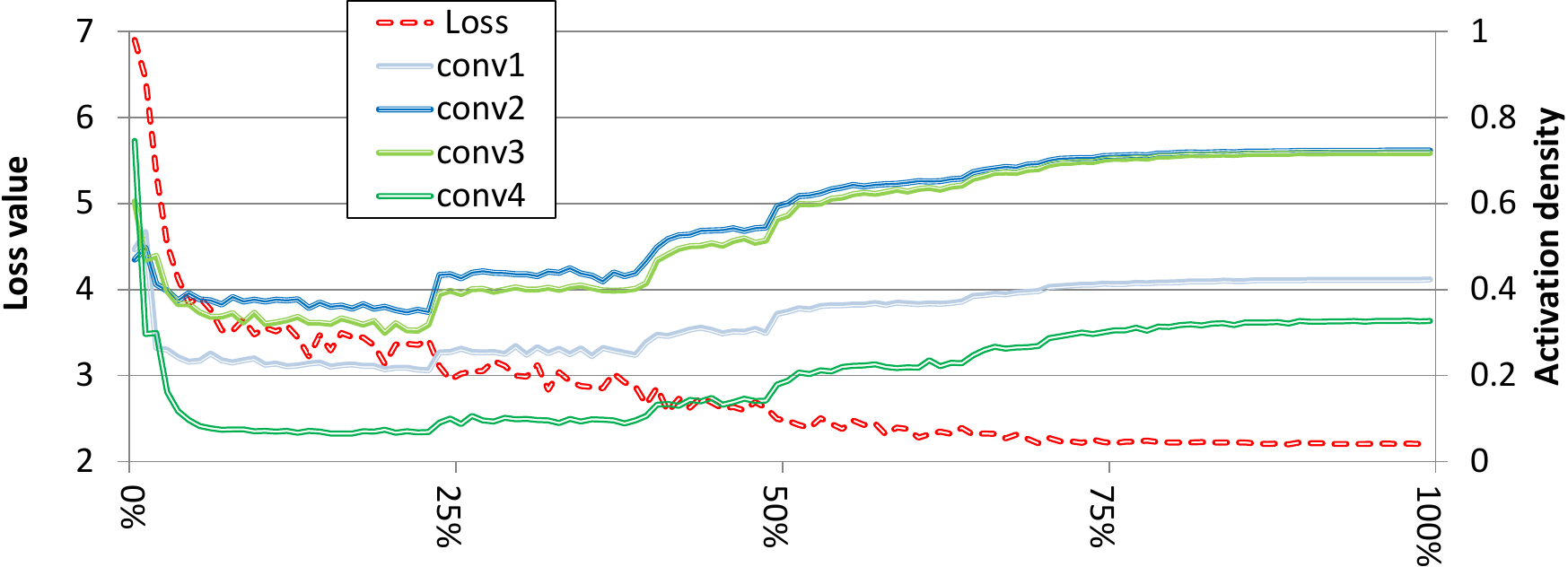}
\caption{Change in loss value (left axis) and per-layer activation density (right axis) as the
network is trained (\texttt{x}-axis).}
\vspace{-1em}
\label{fig:loss_alexnet} 
\end{figure}

\subsection{Effects of Training on Sparsity}
\label{sect:sparsity_summary}

In addition to AlexNet, we examined the sparsity of activations for
larger, deeper, and more recent CNNs, including
OverFeat~\cite{overfeat}, NiN~\cite{nin}, VGG~\cite{vggnet},
SqueezeNet~\cite{squeezenet}, and GoogLeNet~\cite{googlenet}.
Figure~\ref{fig:layer_density} shows that the per-layer sparsity
measurements of these networks are very similar in nature to AlexNet,
reinforcing the observations listed above.  In the six networks that
we study in this paper, we observe an average $62\%$ network-wide
activation sparsity (maximum of $93\%$) across the entire training
periods.  Figure~\ref{fig:loss_alexnet} shows the behavior of AlexNet
as a function of training time, including the loss value computed by
the loss function at the end of the network and the activation
densities of the four convolutional layers. The graph demonstrates
four key observations about the effect of training on per-layer
activation density, as described below.

\begin{itemize}

\item When training begins, activation density drops dramatically for
  all of the layers. This drop correlates with the dramatic
  improvement in the loss function. We believe that this drop in
  density is due to the network quickly adapting from its randomly
  initialized weights to learning what features of the input data are
  not important for classification.

\item In the second regime, the activation density increases, first
  somewhat rapidly and then more slowly. We believe that this increase
  stems from two factors. First, during the middle stages of training,
  the network weights are iteratively optimized to extract features
  that it has previously neglected, gradually improving
  accuracy. Second, a common heuristic in training DNNs is to reduce
  the learning rate, typically multiplying the original learning rate
  by $0.5$ or $0.1$, when the validation accuracy stops increasing
  with the current learning rate~\cite{alexnet,zfnet}.

\item During the final fine-tuning stages of training, the weights are
  already close to their respective optimal points so the effect on
  the overall average activation sparsity is minimal.

\item In general, convolution layers later in the network are sparser
  than earlier ones. Deep networks are known to build up a rich set of
  hierarchical feature extractors across the network layers. For
  instance, Zeiler and Fergus~\cite{zfnet} observed that the first few
  layers of a CNN are generally trained to respond to corners, edges,
  and basic colors that commonly exist across all images in a
  \emph{class-invariant} fashion. However, deeper layers are used to
  detect \emph{class-specific} high-level abstractions such as common
  textures (e.g., mesh patterns), texts, faces of a dog, etc.
  Accordingly, we hypothesize that layers located deep down in the
  network are trained to respond to class-specific features and have
  activations that only respond to a subset of classes, leading to
  high sparsity. In contrast, because layers positioned early in a
  network respond in a class-invariant manner (e.g., activations will
  respond to all the classes that have red-colored regions of an
  image), they are likely to exhibit less sparsity.

\end{itemize}

\section{Compressing DMA Engine}
\label{sect:dma}

To address the performance bottlenecks associated with moving
activation maps between the GPU and CPU memory, we exploit the
sparsity of activation maps to compress them before transferring them
across the PCIe bus.  The overall approach is somewhat similar to
compressing pages prior to moving them to backing storage in a virtual
memory system~\cite{compressed_page_cache}.  Our compressing DMA
engine (\cdma) requires choosing an efficient and effective
compression algorithm, and a mechanism to employ this algorithm to
compress activation maps as they are transferred between the GPU and
CPU memory.

\subsection{Compression Algorithm}

To compress activation maps, we need a simple compression algorithm
which can sustain compression from and decompression into GPU memory
at rates of 100's of GB/sec while saturating PCIe bandwidth with
compressed data.

{\bf Run-length encoding compression.} Early observations of the
sparse activation maps demonstrated a clustering of zero-valued
activations (\fig{fig:layer_density_maps}).  As a result, we investigate a simple
scheme using {\em run-length encoding} (\rle)~\cite{runlengthencoding}
to compress the activation maps.  Run-length encoding is simple to
implement, and is well suited for high-bandwidth compression.  Despite
its simple design, the effectiveness of \rle  highly depends on
the sparsity patterns exhibited in the activation maps as compression
is only effective for consecutive zeros or non-zeros. As a result,
\rle does not offer good compression ratios across all of the
activation layouts (detailed in \sect{sect:compression_eff}).

\begin{figure}[tbp] \centering
\includegraphics[width=0.45\textwidth]{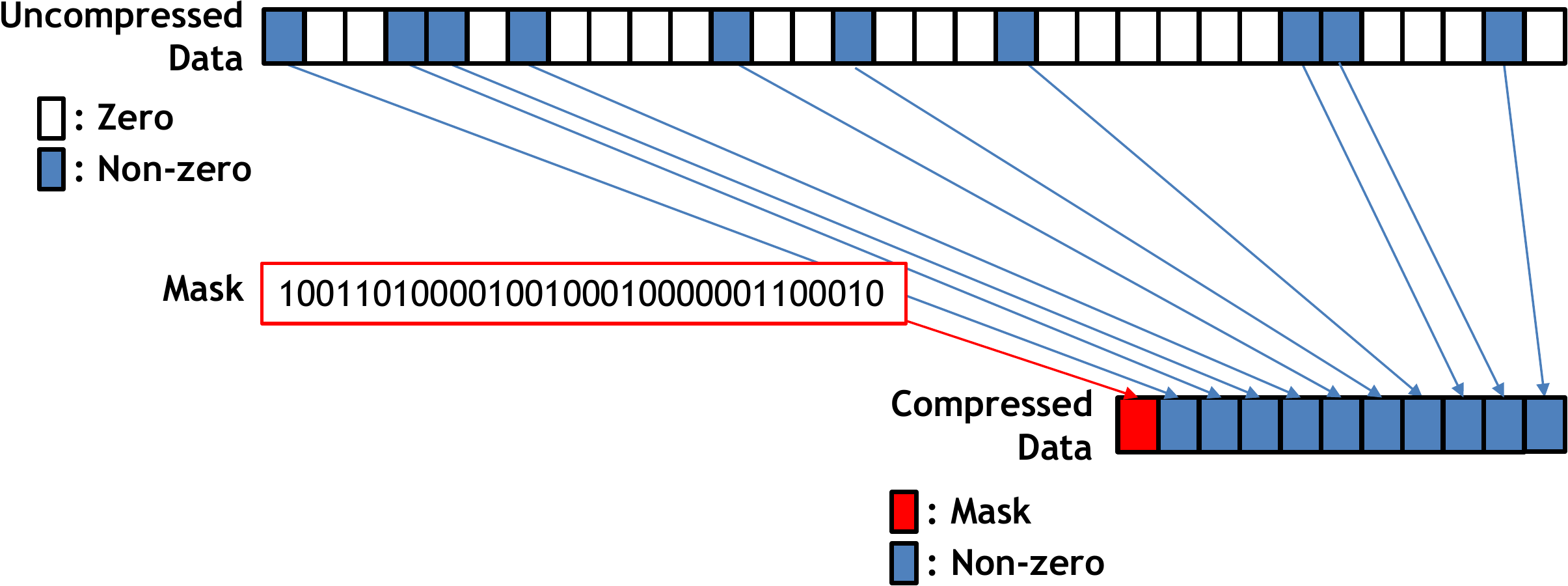}
\caption{Zero-value compression.}
\vspace{-1em}
\label{fig:zvc_compression}
\end{figure}

{\bf Zero-value compression.} As demonstrated in
Section~\ref{sect:characterization}, approximately $50\%$ to $90\%$ of
the activations are zero-valued. We therefore investigate a simple yet
highly effective approach based on {\em Frequent-value
  compression}~\cite{fvc} that is used to compress out the zero-valued
elements.

\fig{fig:zvc_compression} provides a high-level overview of our
\emph{zero-value compression} (\zvc) algorithm which assumes a
compression window sized as $32$ consecutive elements. For every $32$
activation values, a $32$-bit mask is generated with a `$0$' in a
given bit position indicating the value is zero and a `$1$' indicating
a non-zero value.  After this $32$-bit mask is generated, the non-zero
elements are appended.  Thus, $32$ consecutive zero valued activations
can be compressed down to a single $32$-bit all-zero mask ($32\times$
compression ratio).  $32$-consecutive non-zero elements will result in
a $32$-bit all-one mask, followed by the $32$ non-zero activation
values (a $3.1\%$ metadata overhead, $1$-bit per each single
activation value).  If $60\%$ of the total activations are
zero-valued, we would expect an overall compression ratio of
$2.5\times$.
Compared to \rle, the key advantage of
\zvc is that it can compress out zeros equally well regardless of how
the zero values are distributed in the data. Unlike \rle, \zvc works
robustly across all the data layouts of the activation maps.

\zvc can be implemented in high-bandwidth compression hardware in a
straightforward manner.  The hardware implementation complexity is
dominated by the MUXes to gather/scatter the non-zero data elements
to/from the compressed representation and the pop-count/prefix-sum
operation on the mask to determine the offset to the next mask in the
compressed stream. We detail the ZVC DMA engine
  microarchitecture in \sect{sect:dma_Engine} and the area overhead in
  \sect{sect:overhead}. 

{\bf Zlib compression.} The compression scheme used in the popular
{\em gzip} utility is based on the DEFLATE algorithm~\cite{deflate}.
This algorithm has very good performance across a range of data, but
designing a high-throughput hardware to perform the compression is
quite complex.  Dedicated FPGA and ASIC
solutions~\cite{gzip_fpga,gzip_asic} are capable of reaching
approximately $2.5$ GB/sec of throughput.  While processing multiple
streams in parallel with multiple compression engines can improve
throughput, the hardware costs escalate linearly with increased
bandwidth.  Supporting this compression algorithm is impractical when
the system must be capable of compressing 100's of GB/sec of data.
Nonetheless, we include the results using this approach to demonstrate
the upper-bound of the opportunity we may be leaving on the table by
not compressing non-zero data and focusing solely on zero-value
compression.

\begin{figure}[t!] \centering
\includegraphics[width=0.49\textwidth]{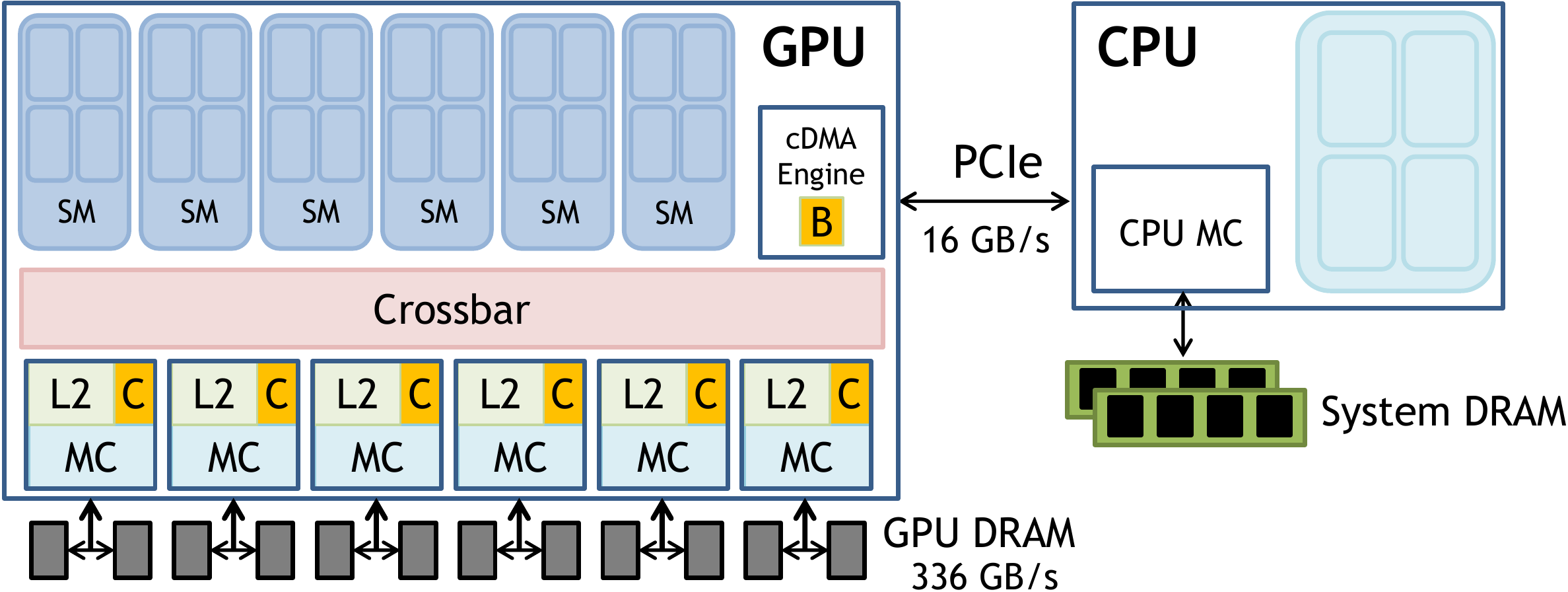}
\caption{\cdma architecture overview. Box ``B'' indicates the
  location of the \cdma buffers whereas boxes labeled ``C''
  indicate the location of (de)compression units.}
\vspace{-1em}
\label{fig:dma_arch}
\end{figure}

\subsection{Compressing DMA Engine Architecture}
\label{sect:dma_Engine}

{\bf Architecture overview.}  \fig{fig:dma_arch} provides an overview of the
\cdma architecture embedded into the memory system of a GPU\@. The additional
hardware includes compression/decompression units adjacent to the GPU memory
controllers (boxes labeled ``C'') and a little more data buffering storage (box
labeled ``B'') in the existing DMA engine at the PCIe interface.  GPUs already
perform compression operations within the memory controllers
today~\cite{Sathish:PACT2012, gpu_compression, nvidia-tegraX1}, but the
compression operations of our \cdma are somewhat backwards compared to existing
systems. The existing compression hardware in the GPU memory controllers
compress data on the way {\em into} the GPU DRAM and decompress on the way out
to the L2 and GPU cores (streaming multiprocessors, denoted as SMs in
\fig{fig:dma_arch}) to save GPU DRAM bandwidth.  Our \cdma architecture
compresses the data coming {\em out} of the GPU DRAM on their way to the DMA
unit, and decompresses data in the other direction. We provide a qualitative 
discussion on how the operation of \cdma can be designed to work in a conventional
way (i.e., compression taking place on its way to the DRAM to save
bandwidth) in \sect{sect:related_discussion}.

An alternative implementation of \cdma would directly add the
compression/decompression units inside the existing DMA unit so that
it compresses the data just before sending it over PCIe, or
alternatively, decompresses the data when received over PCIe from the
CPU\@. One key concern with this design is its effect on the bandwidth
requirements of the GPU on-chip crossbar which connects the memory controllers
to the SMs and DMA engine.

The key design objective of our \cdma engine is to be able to saturate
the PCIe bandwidth to the CPU with compressed data. Accordingly, the
GPU crossbar bandwidth that routes uncompressed data from the L2 to
the DMA engine must be high enough to generate compressed activation
maps at a throughput commensurate to the PCIe link bandwidth. As
detailed in \sect{sect:compression_eff}, the maximum per-layer
compression ratio observed is $13.8\times$. Assuming PCIe (gen3)
with maximum $16$ GB/sec data transfer bandwidth, up to
($16\times13.8$)$=$$220.8$ GB/sec crossbar bandwidth must be
provisioned to fully exploit the potential of sparse compression.
Since the baseline DMA engine need only serve the $16$ GB/sec of PCIe
bandwidth, providing over $200$ GB/sec of crossbar bandwidth to the
DMA engine for the purposes of data offloading is unattractive.

Our \cdma design instead augments the GPU memory controllers with the
(de)compression units to compress the data read from the DRAM \emph{before}
sending it over the crossbar to the DMA. Such design reduces
the bandwidth demand on the crossbar during a compressed DMA operation
back to levels similar to the baseline non-compressing DMA engine.

\begin{figure}[t!] \centering
\subfloat[ZVC compression engine]{
\includegraphics[width=0.45\textwidth]{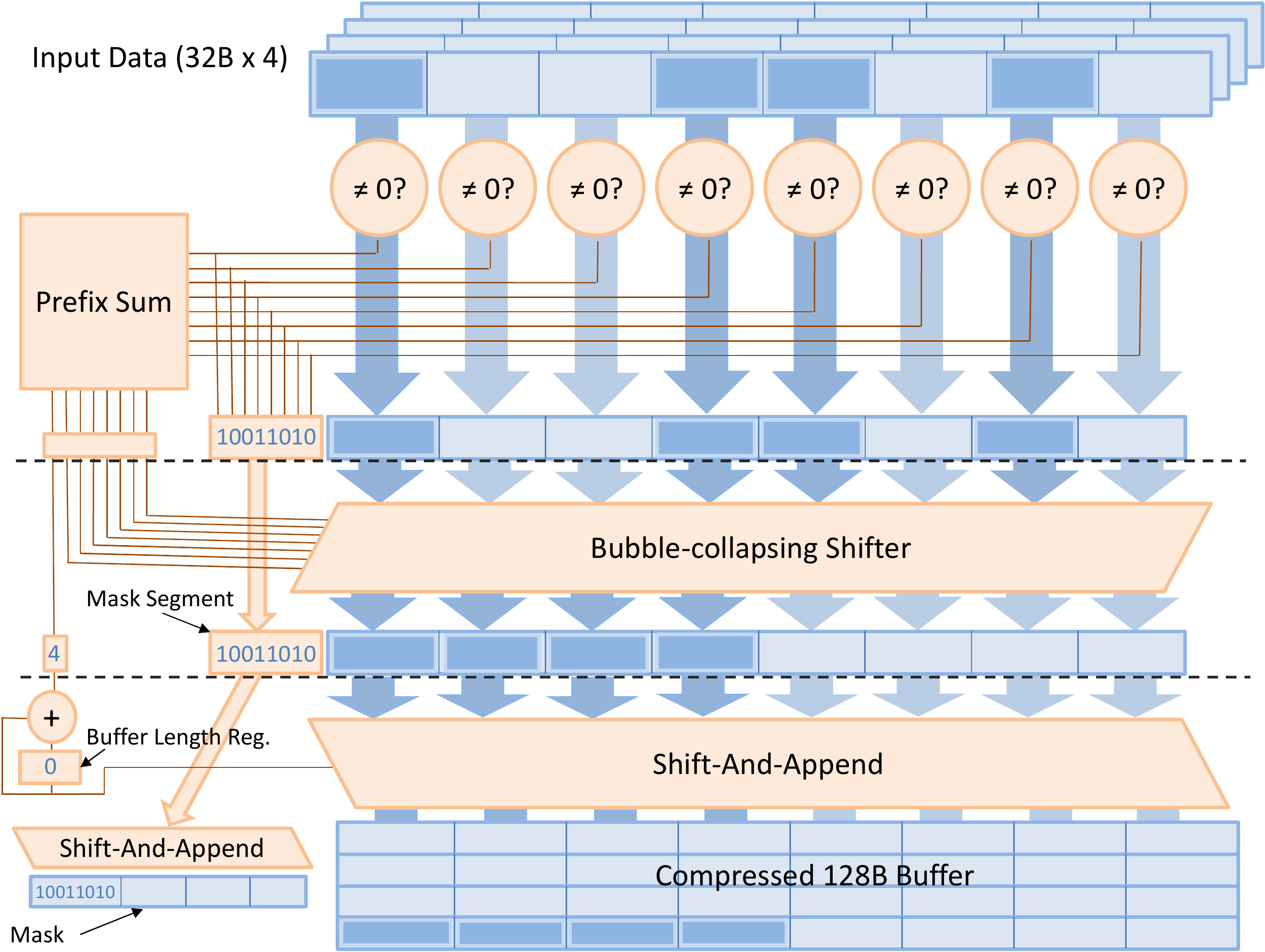}
\label{fig:zvc_compress_uarch}
}
\vspace{0em}
\subfloat[ZVC decompression engine]{
\includegraphics[width=0.45\textwidth]{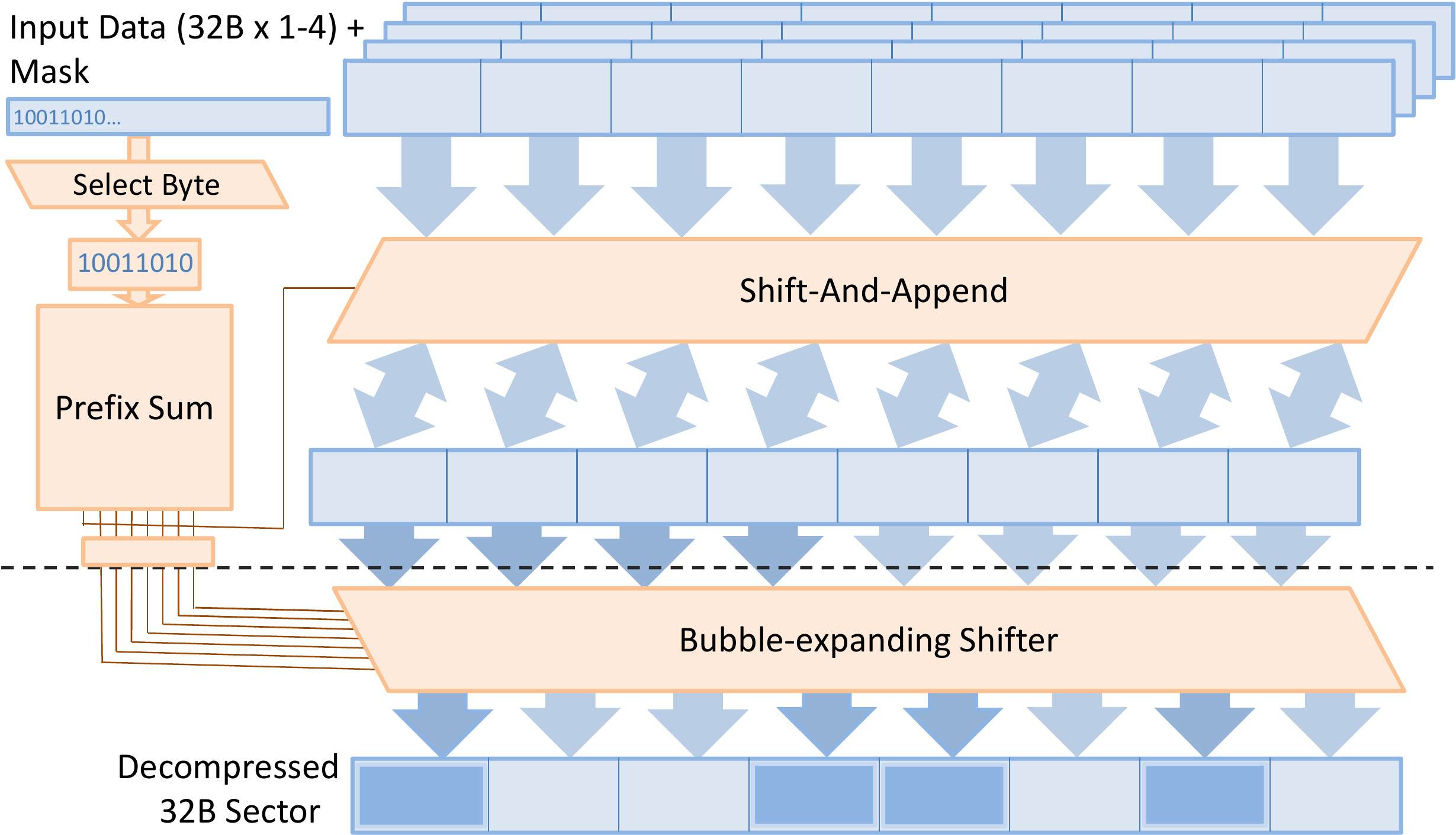}
\label{fig:zvc_decompress_uarch}
}
\caption{ZVC engine microarchitecture.}
\vspace{-1em}
\label{fig:zvc_uarch} 
\end{figure}

{\bf (De)compression engine microarchitecture.}  
\fig{fig:zvc_uarch}(a) shows the microarchitecture of the
compression engine implementing the ZVC algorithm.  This logic
operates each cycle on a 32B (8 word) element which corresponds both
to the internal data-path width in the memory controller and to one
DRAM burst.  On one cycle, these eight words are compared in parallel
to zero, forming the mask bits for these 8 words.  A prefix sum
operation, requiring 11 3-bit adders, is performed on the mask bits to
determine the number of zero-valued words in front of a given word.
In the next pipeline cycle, the non-zero data elements are shifted to
the correct resulting offset using the result of the prefix sum
operation to drive the mux-selects.  The final cycle in the pipeline
steers the resulting zero-compressed data to append it to previous
compressed data in the overall 128-bytes (cache-line sized) window on
which we perform ZVC compression.  The 8-bit mask is also appended to
the mask from previous cycles. The total latency to compress a
128-bytes line is six cycles, four 32B sectors moving through a
three-stage pipeline.

\fig{fig:zvc_uarch}(b) shows the microarchitecture of the ZVC
decompression engine which expands one compressed item into 128 bytes.
The decompression logic also operates on 32B at a time, producing 32B
of decompressed data each cycle.  In the first cycle of the pipeline,
an 8-bit segment of the mask is considered.  A pop-count (number of
ones) of this 8-bit segment of the mask determines the number of words
that will be used in a given 32-byte sector. In parallel, the 8-bit
segment of the mask is evaluated to determine the correct mux-selects
(also a small prefix-sum operation). In the next cycle, the 32-byte
decompressed value is formed by muxing payload values (or a zero) into
the correct locations. The pipeline requires only two additional cycles of
latency to decompress a 128-bytes line, because decompression can
start as soon as the first part of the data arrives from the crossbar.

\subsection{Design Overheads}
\label{sect:overhead}

{\bf (De)compression units.} While we expect that the existing GPU compression
units can be leveraged for \cdma to minimize design overheads, we assume that
the \cdma (de)compression hardware supplements existing hardware for a
conservative area estimate.  Nonetheless, our \cdma unit can allow existing
DRAM compression schemes optimized to minimize DRAM bandwidth to also take
place. We use the FreePDK~\cite{freepdk} $45$\,nm process design kit and scaled
the resulting area using a conservative cell size reduction of $0.46\times$
from $45$\,nm to $28$\,nm.  Assuming a $50\%$ cell area utilization due to the
design being dominated by wires and MUXes, the six (de)compression units are
estimated to incur a $0.31$\,mm$^2$ area overhead.

{\bf Buffer sizing.} The DMA engine must also maintain a buffer large
enough to hold the bandwidth-delay product of the memory sourcing the
data to prevent bubbles in the output stream. As we detail in
\sect{sect:perf}, DNN computations are highly compute-bound so the
required average memory bandwidth is measured at less than $100$
GB/sec, leaving more than ($336-100$)$=$$236$ GB/sec for \cdma to
fetch data without affecting performance. Our experiments show that
provisioning $200$ GB/sec of bandwidth for \cdma reaps most of the
benefits of sparse compression.  As a result, based on a $350$\,ns
latency from the time the DMA engine requests data from GPU memory to
the time it arrives at the DMA engine~\cite{Wong:ISPASS2010} and the
$200$\,GB/sec compression read bandwidth, the DMA engine needs a $70$KB
($200$\,GB/sec$\times$$350$\,ns) buffer, shown as block ``B'' in
\fig{fig:dma_arch}.  It may seem counter-intuitive that \cdma would
need this large a buffer, since it is receiving only compressed requests
at an overall rate of $16$ GB/sec from the crossbar.  The reason why
the buffer needs to be overprovisioned is because the \cdma engine
does not know {\em a priori} which responses will be compressed or
not.  Since it must launch sufficient requests to keep the PCIe bus
busy even with highly-compressed data, a large number of requests will
be in-flight.  In the event that these requests are {\em not}
compressed, the buffer is required to hold the large amount of data
returned until it can be streamed out over the PCIe interface.  This
buffer size is not a significant source of area (approximately $0.21$\,mm$^2$ in $28$\,nm according to CACTI 5.3~\cite{cacti}). Compared to
the $600$\,mm$^2$ of a NVIDIA Titan X chip used for the evaluations in
this paper, the added overheads of (de)compression units and DMA
buffers are negligible.

\subsection{Software Interface}
\label{sect:dma_software_interface}

The compression/decompression features of the DMA engine can be
exposed to the programmer so that it can be adopted within ML
frameworks and other applications.  We envision that the compressed
memory copy operation can be exposed to the SW level using a new {\tt
  cudaMemcpyCompressed()} call that enables the compression (or
decompression) in the DMA engine.  We expect this API will be extended
beyond the typical {\tt cudaMemcpy()} to also return the compressed
size of a region on completion of the copy operation. In our
experimental framework, the {\tt cudaMemcpyCompressed} calls would
easily replace the {\tt cudaMemcpy} calls already deployed in \vdnn.

\section{Evaluation Methodology}
\label{sect:eval}

Architectural exploration of \cdma in cycle-level simulation is
challenging for two primary reasons. First, existing GPU
architecture simulators (e.g., GPGPU-Sim~\cite{gpgpusim} and
GPUOcelot~\cite{gpuocelot}) are not able to execute the cuDNN APIs as
these GPU accelerated library routines are released as pre-compiled
binaries.  Second, a single iteration of training can take up to tens
of seconds even on the fastest GPU, so running cycle-level simulations
on these ML workloads within a reasonable timeframe is likely a
research project on its own.  We therefore take a hybrid approach in
evaluating the effect of \cdma on training performance. Specifically,
we measure DNN applications on a real GPU card while properly
penalizing the system performance based on an analytical model of the
GPU memory subsystem, as summarized below.

{\bf Virtualized DNN.} We faithfully model the \vdnn memory management
policy as described in \cite{rhu:2016:vdnn}, which is interfaced to
the latest version of cuDNN (v5)~\cite{cudnn}.  \vdnn is configured to
offload all the layer's activation maps for memory-scalability and to
maximally stress the PCIe channel. The offload and prefetch operations
to and from CPU memory are initiated using \texttt{cudaMemcpyAsync()};
the memory allocation size is determined by the compression ratio
observed by the \cdma unit, as modeled below.

{\bf Compression pipeline.} We implemented our \cdma compression
pipeline on top of Caffe~\cite{caffe}.  We modified the Caffe Python interface
(\texttt{pycaffe}) to checkpoint the target
network's activations so that they can be fed to our \cdma compression algorithm
to compress and downsize the activation maps for each layer's
offloaded data.  The compressed activation maps are then returned
to the \vdnn memory manager to measure the latency incurred during the
memory copy operation to/from the CPU-side memory across the PCIe bus.

{\bf Effect of \cdma on memory bandwidth.}  Compared to a baseline
implementation of \vdnn, \cdma affects system performance based on the
following two factors.  First, the reduced PCIe traffic helps improve
the performance of \vdnn because the latency to move data in/out of
CPU memory is significantly reduced.  However, to fully saturate the
PCIe link bandwidth and maximize the benefits of DNN virtualization,
the compressed activations must be generated at a throughput
commensurate to the PCIe transfer bandwidth.  Thus the second issue is
that the average DRAM bandwidth utilization of \cdma can exceed that
of \vdnn by a factor of $2.6\times$ (\sect{sect:compression_eff}),
potentially interfering with the cuDNN computation and decreasing
performance.

State-of-the-art DNN libraries refactor the convolution operations
into a dense matrix-multiplication operation for GPU
acceleration~\cite{chetlur:2014:cudnn}.  This approach allows the DNN
computation to be completely \emph{compute-bound} with high cache hit
rates and low average DRAM bandwidth utilization. Using the NVIDIA
CUDA profiler (\texttt{nvprof}), we observed less than an average of
100 GB/sec of off-chip memory bandwidth utilization across all six
networks. These numbers are consistent with the results reported by
Rhu et al.~\cite{rhu:2016:vdnn}, leaving more than an average
$336-100=236$ GB/sec of memory bandwidth available for our \cdma
engine to fetch activation maps from the GPU memory without affecting
the throughput of DNN computations using cuDNN\@.

As we are not able to model \cdma inside an existing GPU,
evaluating the performance of cuDNN with both \vdnn and \cdma in
silicon is impractical. Nonetheless, as long as the GPU memory
bandwidth consumption of \cdma (i.e., a given layer's compression
ratio$\times$PCIe transfer bandwidth, denoted as \texttt{COMP\_BW}
below) is smaller than the available $236$ GB/sec of DRAM bandwidth
(\texttt{DRAM\_BW}), the compressed activations can be generated at a
high enough rate to fully saturate the PCIe bandwidth while not
affecting the performance of baseline cuDNN.

To model the bandwidth limitations on \cdma performance, we restrict
the memory bandwidth consumption of \cdma to never exceed the $236$
GB/sec leftover bandwidth of Titan X\@.  For the few layers that do
require a DRAM bandwidth higher than $236$ GB/sec (i.e., layers with
compression ratio$\times$PCIe transfer bandwidth higher than
\texttt{DRAM\_BW}), we assume that the compressed activations are not
generated at a fast enough rate to saturate the PCIe channel. In other
words, when evaluating system performance of \vdnn, we \emph{increase}
the latency incurred when offloading the compressed activations by a
factor of (\texttt{COMP\_BW}/\texttt{DRAM\_BW}), modeled in an
existing GPU by inflating the volume of data transferred over the PCIe
interface. For a conservative evaluation, we set the \texttt{COMP\_BW}
value, the maximum memory bandwidth \cdma is allowed to
consume\footnote{Even though the peak memory bandwidth consumption of
  \cdma can be on the order of $200$ GB/sec, the \emph{average} memory
  bandwidth usage will not exceed $16{\times}2.6=41.3$ GB/sec,
  which is the PCIe bandwidth$\times$average network-wide compression
  ratio of $2.6$.}, to $200$ GB/sec.

{\bf GPU node topology.} Our DNN training platform contains an Intel
i7-5930K CPU with $64$ GB of DDR4 memory communicating with an NVIDIA
Titan X (Maxwell) containing $12$ GB of GDDR5 memory with a maximum of
$336$ GB/sec bandwidth~\cite{titan_x}.  The PCIe switch (gen3)
provides a maximum of $16$ GB/sec of data transfer bandwidth.

{\bf Training methodology.} All networks are trained using stochastic
gradient descent (SGD) with an initial learning rate of $0.01$. We
manually reduce the learning rate by factor of $0.1$ or $0.5$,
choosing the value that provides higher improvements in validation
accuracy when the validation error plateaus.  Dropout~\cite{dropout}
is employed for the fully-connected layers with a rate of $0.5$. We
terminate the training process when the validation accuracy does not
improve further beyond a learning rate smaller than $1\e{-5}$.  All of
our compression algorithms are \emph{lossless} and affects neither the
functionality nor the algorithmic nature of SGD\@.

{\bf Networks evaluated.} We study DNNs that show state-of-the-art
performance in ImageNet~\cite{imagenet}: AlexNet~\cite{alexnet},
OverFeat~\cite{overfeat}, NiN~\cite{nin}, VGG~\cite{vggnet},
SqueezeNet~\cite{squeezenet}, and GoogLeNet~\cite{googlenet}.  We
configure these networks based on the \texttt{.prototxt} files
available at Caffe Zoo~\cite{caffe} or ones available at the original
authors' websites~\cite{nin,squeezenet}. \tab{tab:networks} summarizes
each network's fully trained top-1/top-5 classification accuracy, the
minibatch sizes used for training, and the total number of training
iterations taken to reach its final trained model.

\begin{table}[t!]
  \centering
  \caption{Networks and trained model accuracy. }
  \vspace*{0.1in}
  \begin{tabular}{|c|c|c|c|}
    \hline
    \textbf{Network} & \textbf{Top-1/Top-5 ($\%$)} & \textbf{Batch} & \textbf{Trained iter.} \\
    \hline
    \hline
    AlexNet     & $53.1$ / $75.1$ & 256 & $226$K \\
    \hline                              
    OverFeat    & $52.8$ / $76.4$ & 256 & $130$K \\
    \hline                              
    NiN         & $55.9$ / $78.7$ & 128 & $300$K \\
    \hline                              
    VGG         & $56.5$ / $82.9$ & 128 & $130$K \\
    \hline                              
    SqueezeNet  & $53.1$ / $77.8$ & 512 & $82$K \\
    \hline                              
    GoogLeNet   & $56.1$ / $83.4$ & 256 & $212$K \\
    \hline
  \end{tabular}
  \label{tab:networks}
\end{table}

\begin{figure*}[t!] \centering
\includegraphics[width=\textwidth]{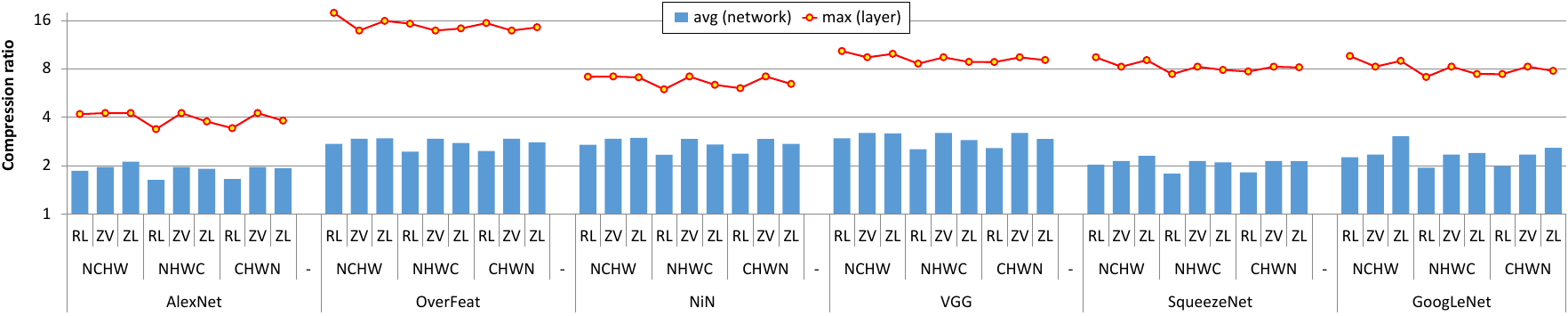}
\caption{The average and maximum compression ratio for a given compression
algorithm for different activation data layouts. The average compression ratio is weighted
by the size of the activation maps that are being offloaded to the CPU so that
it properly reflects the reduction in PCIe traffic. The compression window is
configured as $4$ KB for all three compression schemes. The y-axis is
plotted on a $log_2$ scale.} 
\label{fig:compression_rate_sensitivity_data_layout}
\end{figure*}


\section{Results} 
\label{sect:results}

This section evaluates the efficiency of \cdma compression, the
savings in PCIe traffic, and the effect of \cdma on energy-efficiency
and performance.  The three compression algorithms discussed in
\sect{sect:dma} are denoted as \texttt{RL} (run-length encoding),
\texttt{ZV} (zero-value compression), and \texttt{ZL} (zlib) in all of
the figures discussed in this section.  \vdnn is evaluated with the
memory management policy that provides memory-scalability, which
offloads all activation maps. We also established an \emph{oracular}
baseline (\texttt{orac}) that completely removes the PCIe bottleneck
by having the offload/prefetch latencies always be hidden inside the
DNN computation when measuring performance.

\subsection{Compression Efficiency}
\label{sect:compression_eff}

\fig{fig:compression_rate_sensitivity_data_layout} shows the the
\emph{maximum} per-layer compression ratio achieved across a given
network and the \emph{average} network-wide compression ratio for each
of the three compression algorithms and three data layouts. While the
results presented in this section assume a 4KB compression window, we
also studied window sizes of up to 64KB and found that our results did
not change much.

The maximum per-layer compression ratio determines how much DRAM
bandwidth \cdma must provision to generate the compressed activations
at a high enough rate to fully saturate the PCIe bandwidth.  The
average network-wide compression ratio reflects the reduction in PCIe
traffic provided by \cdma\@.  Overall, our \zvc algorithm provides the
best average compression ratio across all the networks across the
three data layouts (average $2.6\times$).  Despite its simple design,
the efficiency of \zvc is decoupled from the sparsity patterns in the
activation maps and provides the same compression ratio regardless of
how the activations are arranged in GPU memory. \zlib shows the
highest average compression ratio of $2.76\times$ with \nchw but falls
behind \zvc for all but GoogLeNet with \nhwc and \chwn.  Similar to
\zlib, \rle performs best with \nchw but provides the worst
compression with high sensitivity to the underlying data layouts. As
mentioned in \sect{sect:dma}, \zlib and \rle prefer \nchw because this
layout makes it more likely to have the activation sparsity in a
spatially clustered manner. In the rest of this paper, we assume the
\nchw layout for both brevity and for a conservative evaluation of
\zvc as both \rle and \zlib perform best with \nchw.

\begin{figure*}[t!] \centering
\includegraphics[width=0.98\textwidth]{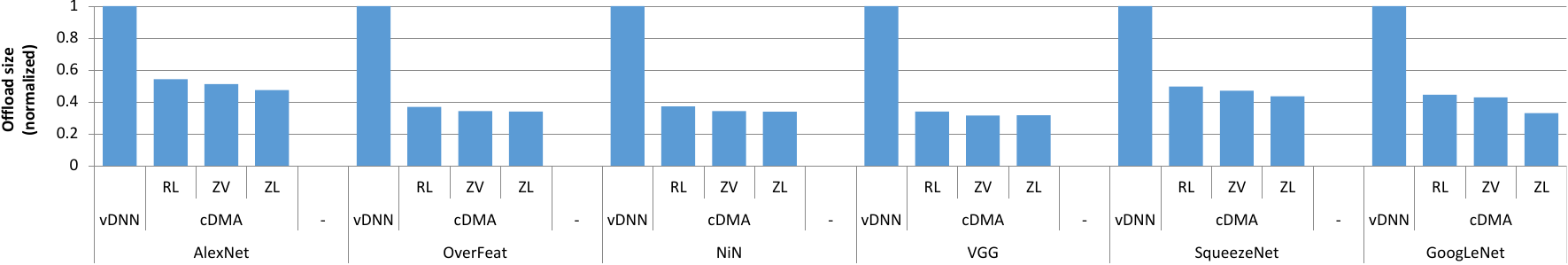}
\caption{
The size of the activation maps offloaded to CPU memory (normalized to \vdnn).}
\label{fig:cpu_malloc_size}
\end{figure*}

\fig{fig:cpu_malloc_size} shows the reduction in the size of the
activations offloaded to the CPU, which directly translates into PCIe
traffic reduction.  Although the sophisticated \zlib algorithm
provides a further 30\% reduction in PCIe traffic for GoogLeNet
(over \zvc), the average traffic reduction across all
six networks is only 3\% compared to \zvc.

\subsection{Performance}
\label{sect:perf}

\begin{figure*}[t!] \centering
\includegraphics[width=0.98\textwidth]{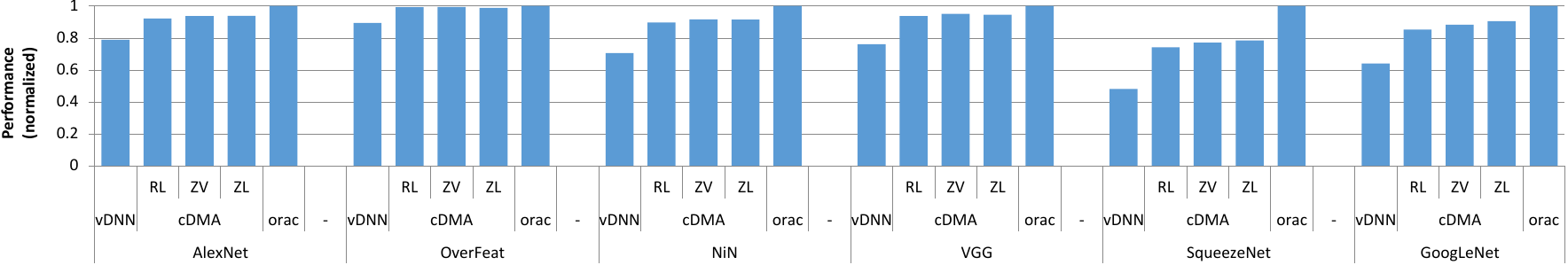}
\caption{Overall performance (normalized to oracle baseline).} 
\label{fig:cdnn_perf} 
\end{figure*}

\fig{fig:cdnn_perf} summarizes the performance of \cdma compared to
\vdnn and the oracular baseline. While \zlib provides the highest
compression ratio for SqueezeNet and GoogLeNet ($8\%$ and $30\%$
higher than \zvc), the resulting performance improvements are
marginal, providing an average $0.7\%$ speedup over \zvc (maximum
$2.2\%$ for GoogLeNet).  The meager performance advantage of \zlib is
twofold: (1) a significant fraction of the offloading latency is
already being hidden by the DNN forward and backward propagation
operations, and (2) the higher compression ratios \zlib achieves are
for layers of which \rle and \zvc already are able to mostly hide the
offloading latencies.  Because of its simple compression algorithm and
robustness across different data layouts, \zvc is the best option for
DNN virtualization.

\subsection{Energy Efficiency}
\label{sect:energy}

The current CUDA software stack does not provide users the ability to
change the DRAM read bandwidth or PCIe transfer bandwidth, making it
difficult to precisely measure the effect of \cdma on
energy-efficiency. Instead, we provide a qualitative comparison of
\cdma's energy-efficiency versus \vdnn. The primary energy overheads
\cdma imposes on \vdnn are (1) the average 2.6$\times$ increase in
DRAM read bandwidth, corresponding to \zvc's average network-wide
compression ratio, for fetching the activations from the
DRAM for \cdma compression; and (2) the (de)compression units and
buffers augmented inside the GPU\@. Based on the examination of
overheads of \cdma in \sect{sect:overhead}, we expect the energy costs
for the additional compression logic and its buffers to be negligible
as \cdma primarily leverages (de)compression units already existing in
GPU memory controllers.

Also, while \vdnn's offload/prefetch operations does incur 1--7\%
power overhead, as measured with \texttt{nvprof}, \cdma's
average $2.6\times$ reduction in PCIe traffic will significantly
reduce the energy-consumption on the PCIe link as well as in the CPU
memory subsystem. When accounting for the average $32\%$ performance
improvements (maximum $61\%$) provided by \cdma, we expect the overall
energy consumption to be significantly improved compared to \vdnn.

\section{Related Work}
\label{sect:related}

DNNs are generally over-parameterized, stemming from a significant
redundancy in the way the parameters are used to represent the model
they are trying to approximate. As a result, there have been series of
proposals that try to reduce the memory usage of DNNs by
alleviating network redundancy.  Vanhoucke et al.~\cite{vanhoucke:2011:snn} explored quantization in activations by
fixing the data type to a $8$-bit integer as opposed to $32$-bit
floating point.  Gong et al.~\cite{gong:2014:compression} proposed
vector quantization methods for compressing the weights of DNNs\@.
Network pruning strategies have also been explored extensively by
prior
literature~\cite{hanson:1989:mnc,lecun:1990:brain_damage,hassibi:1993:brain_surgeon,song:2015:pruning,song:2015:compression}\@.
Pruning helps reduce the memory allocated for model weights by
removing redundant network connections that satisfy a given pruning
criteria.  These proposals provide limited opportunity for saving
memory usage as weights only account for a small fraction of overall
memory allocations needed for training DNNs\@.

The Network-in-Network~\cite{nin} is an approach that tries to
increase the representational power of DNNs by exploiting $1\times1$
convolutional layers. GoogLeNet~\cite{googlenet} and
SqueezeNet~\cite{squeezenet} extensively use these $1\times1$
layers for reducing the dimension of each layer's output
activations. This approach helps remove the computational
bottlenecks while at the same time reducing each layer's memory
requirements.

A series of accelerator designs have also been proposed for
CNNs~\cite{diannao,dadiannao,eyeriss,song:2015:eie,eyeriss_isca,redeye,minerva,dnn_pim_reram,isacc,cnvlutin,shidiannao}\@.
Although these ASIC designs improve the energy-efficiency of CNNs,
these prior proposals focus on inference while \cdma focuses on
training. More importantly, none of these prior works address the
communication bottleneck that arise due to DNN memory virtualization.


\section{Discussion and Future Work}
\label{sect:related_discussion}

{\bf Future CPU-GPU interconnects.}  NVLINK is NVIDIA's proprietary,
high-bandwidth interconnect that enables fast communication between
the GPU and CPU, and between GPUs~\cite{nvlink}.  When coupled with
the IBM Power systems~\cite{ibm_power}, the communication bandwidth
between the CPU-GPU can be up to $80$ GB/sec, greatly alleviating the
communication bottleneck of virtualized DNNs\@. Nonetheless, with a
multi-GPU DNN platform~\cite{alex_weird_trick,dgx_1} where $4$ to $8$
GPUs share the same communication channel, the bandwidth allocated per
each single GPU is still 10--20 GB/sec, similar to PCIe (gen3). As a
result, reducing the offloading traffic between CPU and GPU is still
extremely important.  In addition, because NVLINK is not compatible
with legacy x86 CPUs, the lower bandwidth PCIe interface between CPU
and GPU continues to motivate \cdma for future x86+GPU systems.

{\bf Compression for GPU footprint reduction.} While our \cdma engine
greatly reduces the PCIe traffic and CPU-side memory footprint, the
amount of memory allocated inside the GPU is the same as the baseline
\vdnn.  To reduce GPU DRAM bandwith and memory capacity requirements,
the compression engine inside the GPU's memory controllers could
compress and store the activation maps inside the GPU's DRAM\@.
Implementing this optimization involves developing efficient memory
addressing schemes that allow the memory controller to retrieve the
data in its original, uncompressed form without disturbing overall
performance and energy-efficiency. This future work is beyond the
scope of this paper.

\section{Conclusion}
\label{sect:conclusion}

Previous DNN virtualization solutions can incur significant
performance overheads when the communication channel between the CPU
and GPU is bottlenecked. We introduce a general purpose compressing
DMA engine that can be used for high-performance DNN virtualization.
Our proposal exploits the unique characteristics of DNNs to develop a
cost-efficient compression algorithm, offering an average $2.6\times$
(maximum $13.8\times$) savings in data movement on the CPU--GPU
communication link. Overall, our \cdma engine improves the performance
of virtualized DNNs by an average $32\%$ (maximum $61\%$) with a
modest implementation overhead and can easily be adopted into existing
ML frameworks.

\bibliographystyle{ieeetr}
\bibliography{main}

\begin{thebibliography}{10}

\bibitem{alexnet}
A.~Krizhevsky, I.~Sutskever, and G.~Hinton, ``{ImageNet Classification with
  Deep Convolutional Neural Networks},'' in {\em Proceedings of the
  International Conference on Neural Information Processing Systems (NIPS)},
  December 2012.

\bibitem{graves:2005:fpc}
A.~Graves and J.~Schmidhuber, ``{Framewise Phoneme Classification With
  Bidirectional LSTM and Other Neural Network Architectures},'' {\em Neural
  Networks}, pp.~602--610, 2005.

\bibitem{collobert:2011:nlp_from_scratch}
R.~Collobert, J.~Weston, L.~Bottou, M.~Karlen, K.~Kavukcuoglu, and P.~Kuksa,
  ``{Natural Language Processing (Almost) From Scratch}.''
  \url{https://arxiv.org/abs/1103.0398}, 2011.

\bibitem{caffe}
Caffe. \url{http://caffe.berkeleyvision.org}, 2016.

\bibitem{torch}
Torch. \url{http://torch.ch}, 2016.

\bibitem{theano}
Theano. \url{http://deeplearning.net/tutorial}, 2016.

\bibitem{tensorflow}
Tensorflow. \url{https://www.tensorflow.org}, 2016.

\bibitem{neon}
Nervana. \url{https://github.com/NervanaSystems/neon}, 2016.

\bibitem{mxnet}
T.~Chen, M.~Li, Y.~Li, M.~Lin, N.~Wang, M.~Wang, T.~Xiao, B.~Xu, C.~Zhang, and
  Z.~Zhang, ``{MXNet: A Flexible and Efficient Machine Learning Library for
  Heterogeneous Distributed Systems},'' in {\em Proceedings of the Workshop on
  Machine Learning Systems}, December 2015.

\bibitem{cntk}
Microsoft. \url{https://github.com/Microsoft/CNTK}, 2016.

\bibitem{cudnn}
NVIDIA, ``{cuDNN: GPU Accelerated Deep Learning}.''
  \url{https://developer.nvidia.com/cudnn}, 2016.

\bibitem{rhu:2016:vdnn}
M.~Rhu, N.~Gimelshein, J.~Clemons, A.~Zulfiqar, and S.~W. Keckler, ``{vDNN:
  Virtualized Deep Neural Networks for Scalable, Memory-Efficient Neural
  Network Design},'' in {\em Proceedings of the International Symposium on
  Microarchitecture (MICRO)}, October 2016.

\bibitem{vggnet}
K.~Simonyan and A.~Zisserman, ``{Very Deep Convolutional Networks for
  Large-Scale Image Recognition}.'' \url{https://arxiv.org/abs/1409.1556}, May
  2015.

\bibitem{googlenet}
C.~Szegedy, W.~Liu, Y.~Jia, P.~Sermanet, S.~Reed, D.~Anguelov, D.~Erhan,
  V.~Vanhoucke, and A.~Rabinovich, ``{Going Deeper with Convolutions},'' in
  {\em Proceedings of the Conference on Computer Vision and Pattern Recognition
  (CVPR)}, June 2015.

\bibitem{nn_stochastic_depth}
G.~Huang, Y.~Sun, Z.~Liu, D.~Sedra, and K.~Weinberger, ``{Deep Networks with
  Stochastic Depth}.'' \url{https://arxiv.org/abs/1603.09382}, 2016.

\bibitem{lecun_gd}
Y.~LeCun, L.~Bottou, Y.~Bengio, and P.~Haffner, ``{Gradient-Based Learning
  Applied to Document Recognition},'' {\em Proceedings of the IEEE}, vol.~86,
  pp.~2278--2324, November 1998.

\bibitem{chetlur:2014:cudnn}
S.~Chetlur, C.~Woolley, P.~Vandermersch, J.~Cohen, J.~Tran, B.~Catanzaro, and
  E.~Shelhamer, ``{cuDNN: Efficient Primitives for Deep Learning},'' in {\em
  Proceedings of the International Conference on Neural Information Processing
  Systems (NIPS)}, December 2014.

\bibitem{cuda_convnet}
A.~Krizhevsky, ``{cuda-convnet}.''
  \url{https://code.google.com/p/cuda-convnet/}, 2012.

\bibitem{gpu_tlb}
B.~Pichai, L.~Hsu, and A.~Bhattacharjee, ``{Architectural Support for Address
  Translation on GPUs: Designing Memory Management Units for CPU/GPUs with
  Unified Address Spaces},'' in {\em Proceedings of the International
  Conference on Architectural Support for Programming Languages and Operation
  Systems (ASPLOS)}, March 2014.

\bibitem{gpu_x86_at}
J.~Power, M.~Hill, and D.~Wood, ``{Supporting x86-64 Address Translation for
  100s of GPU Lanes},'' in {\em Proceedings of the International Symposium on
  High-Performance Computer Architecture (HPCA)}, February 2014.

\bibitem{gpu_paging}
T.~Zheng, D.~Nellans, A.~Zulfiqar, M.~Stephenson, and S.~W. Keckler, ``{Toward
  High-Performance Paged-Memory for GPUs},'' in {\em Proceedings of the
  International Symposium on High-Performance Computer Architecture (HPCA)},
  March 2016.

\bibitem{xu:2015:eval_relu}
B.~Xu, N.~Wang, T.~Chen, and M.~Li, ``{Empirical Evaluation of Rectified
  Activations in Convolutional Network}.''
  \url{https://arxiv.org/abs/1505.00853}, 2015.

\bibitem{sun:2015:face_sparse}
Y.~Sun, X.~Wang, and X.~Tang, ``{Deeply Learned Face Representations Are
  Sparse, Selective, and Robust},'' in {\em Proceedings of the Conference on
  Computer Vision and Pattern Recognition (CVPR)}, June 2015.

\bibitem{cnvlutin}
J.~Albericio, P.~Judd, T.~Hetherington, T.~Aamodt, N.~E. Jerger, and
  A.~Moshovos, ``{Cnvlutin: Ineffectual-Neuron-Free Deep Convolutional Neural
  Network Computing},'' in {\em Proceedings of the International Symposium on
  Computer Architecture (ISCA)}, June 2016.

\bibitem{imagenet}
ImageNet. \url{http://image-net.org}, 2016.

\bibitem{deepspeech_1}
A.~Hannun, C.~Case, J.~Casper, B.~Catanzaro, G.~Diamos, E.~Elsen, R.~Prenger,
  S.~Satheesh, S.~Sengupta, A.~Coates, and A.~Y. Ng, ``{Deep Speech: Scaling Up
  End-To-End Speech Recognition}.'' \url{https://arxiv.org/abs/1412.5567},
  2014.

\bibitem{deepspeech_2}
D.~Amodei, R.~Anubhai, E.~Battenberg, C.~Case, J.~Casper, B.~Catanzaro,
  J.~Chen, M.~Chrzanowski, A.~Coates, G.~Diamos, E.~Elsen, J.~Engel, L.~Fan,
  C.~Fougner, T.~Han, A.~Hannun, B.~Jun, P.~LeGresley, L.~Lin, S.~Narang,
  A.~Ng, S.~Ozair, R.~Prenger, J.~Raiman, S.~Satheesh, D.~Seetapun,
  S.~Sengupta, Y.~Wang, Z.~Wang, C.~Wang, B.~Xiao, D.~Yogatama, J.~Zhan, and
  Z.~Zhu, ``{Deep Speech 2: End-To-En Speech Recognition in English and
  Mandarin}.'' \url{https://arxiv.org/abs/1512.02595}, 2015.

\bibitem{persistent_rnn}
G.~Diamos, S.~Sengupta, B.~Catanzaro, M.~Chrzanowski, A.~Coates, E.~Elsen,
  J.~Engel, A.~Hannun, and S.~Satheesh, ``{Persistent RNNs: Stashing Recurrent
  Weights On-Chip},'' in {\em {Proceedings of the International Conference on
  Machine Learning (ICML)}}, June 2016.

\bibitem{lstm}
S.~Hochreiter and J.~Schmidhuber, ``{Long Short Term Memory},'' {\em Neural
  Computation}, vol.~9, pp.~1735--1780, November 1997.

\bibitem{gru}
J.~Chung, C.~Gulcehre, K.~Cho, and Y.~Bengio, ``{Gated Feedback Recurrent
  Neural Networks}.'' \url{https://arxiv.org/abs/1502.02367}, 2015.

\bibitem{overfeat}
P.~Sermanet, D.~Eigen, X.~Zhang, M.~Mathieu, R.~Fergus, and Y.~LeCun,
  ``{OverFeat: Integrated Recognition, Localization and Detection using
  Convolutional Networks}.'' \url{https://arxiv.org/abs/1312.6229}, 2013.

\bibitem{nin}
M.~Lin, Q.~Chen, and S.~Yan, ``{Network in Network}.''
  \url{https://arxiv.org/abs/1312.4400}, 2013.

\bibitem{squeezenet}
F.~Iandola, S.~Han, M.~Moskewicz, K.~Ashraf, W.~J. Dally, and K.~Keutzer,
  ``{SqueezeNet: AlexNet-level Accuracy with 50x Fewer Parameters and $<$0.5MB
  Model Size}.'' \url{https://arxiv.org/abs/1602.07360}, 2016.

\bibitem{zfnet}
M.~Zeiler and R.~Fergus, ``{Visualizing and Understanding Convolutional
  Networks}.'' \url{https://arxiv.org/abs/1311.2901}, 2013.

\bibitem{compressed_page_cache}
P.~Wilson, S.~Kaplan, and Y.~Smaragdakis, ``{The Case for Compressed Cache in
  Virtual Memory Systems},'' in {\em Proceedings of USENIX}, June 1999.

\bibitem{runlengthencoding}
A.~Robinson and C.~Cherry, ``{Results of a Prototype Television Bandwidth
  Compression Scheme},'' {\em Proceedings of the IEEE}, vol.~55, pp.~356--364,
  March 1967.

\bibitem{fvc}
Y.~Zhang, J.~Yang, and R.~Gupta, ``{Frequent Value Locality and Value-centric
  Data Cache Design},'' in {\em Proceedings of the International Conference on
  Architectural Support for Programming Languages and Operation Systems
  (ASPLOS)}, November 2000.

\bibitem{deflate}
J.~Gailly and M.~Adler, ``{The gzip Home Page}.''
  \url{http://www.gzip.org/algorithm.txt}.

\bibitem{gzip_fpga}
M.~Abdelfattah, A.~Hagiescu, and D.~Singh, ``{Gzip on a Chip: High Performance
  Lossless Data Compression on FPGAs Using OpenCL},'' in {\em Proceedings of
  the International Workshop on OpenCL}, May 2014.

\bibitem{gzip_asic}
J.~D. Deaton and A.~Bacon, ``{White Paper: Smashing Big Data Costs with GZIP
  Hardware}.'' \url{http://www.aha.com/Uploads/GZIP_Benefits_Whitepaper11.pdf}.

\bibitem{Sathish:PACT2012}
V.~Sathish, M.~Schulte, and N.~Kim, ``{Lossless and Lossy Memory I/O Link
  Compression for Improving Performance of GPGPU Workloads},'' in {\em
  Proceedings of the International Conference on Parallel Architectures and
  Compilation Techniques (PACT)}, September 2012.

\bibitem{gpu_compression}
G.~Pekhimenko, E.~Bolotin, N.~Vijaykumar, O.~Mutlu, T.~C. Mowry, and S.~W.
  Keckler, ``{A Case for Toggle-Aware Compression for GPU Systems},'' in {\em
  Proceedings of the International Symposium on High-Performance Computer
  Architecture (HPCA)}, March 2016.

\bibitem{nvidia-tegraX1}
``{NVIDIA Tegra X1: NVIDIA's New Mobile Superchip}.''
  \url{http://international.download.nvidia.com/pdf/tegra/Tegra-X1-whitepaper-v1.0.pdf},
  2015.

\bibitem{freepdk}
NCSU, ``{FreePDK Process Design Kit}.''
  \url{http://www.eda.ncsu.edu/wiki/FreePDK}, 2016.

\bibitem{Wong:ISPASS2010}
H.~Wong, M.~M. Papadopoulou, M.~Sadooghi-Alvandi, and A.~Moshovos,
  ``{Demystifying GPU Microarchitecture Through Microbenchmarking},'' in {\em
  {Proceedings of the International Symposium on Performance Analysis of
  Systems Software (ISPASS)}}, March 2010.

\bibitem{cacti}
{HP Labs}, ``{CACTI: An Integrated Cache and Memory Access Time, Cycle Time,
  Area, Leakage, and Dynamic Power Model}.''
  \url{http://www.hpl.hp.com/research/cacti/}, 2016.

\bibitem{gpgpusim}
``{GPGPU-Sim}.'' \url{http://www.gpgpu-sim.org/}, 2016.

\bibitem{gpuocelot}
``{GPU Ocelot: A Dynamic Compilation Framework for GPU Computing}.''
  \url{http://gpuocelot.gatech.edu/}, 2016.

\bibitem{titan_x}
NVIDIA, ``{GeForce GTX Titan X (Maxwell)}.''
  \url{http://www.geforce.com/hardware/desktop-gpus/geforce-gtx-titan-x}, 2015.

\bibitem{dropout}
N.~Srivastava, G.~Hinton, A.~Krizhevsky, I.~Sutskever, and R.~Salakhutdinov,
  ``{Dropout: A Simple Way to Prevent Neural Networks from Overfitting},'' {\em
  Journal of Machine Learning Research}, vol.~15, pp.~1929--1958, June 2014.

\bibitem{vanhoucke:2011:snn}
V.~Vanhoucke, A.~Senior, and M.~Mao, ``{Improving the Speed of Neural Networks
  on CPUs},'' in {\em Deep Learning and Unsupervised Feature Learning
  Workshop}, December 2011.

\bibitem{gong:2014:compression}
Y.~Gong, L.~Liu, M.~Yang, and L.~Bourdev, ``{Compressing Deep Convolutional
  Networks Using Vector Quantization}.'' \url{https://arxiv.org/abs/1412.6115},
  2014.

\bibitem{hanson:1989:mnc}
S.~Hanson and L.~Pratt, ``{Comparing Biases for Minimal Network Construction
  with Back-propagation},'' in {\em Proceedings of the International Conference
  on Neural Information Processing Systems (NIPS)}, November 1989.

\bibitem{lecun:1990:brain_damage}
Y.~LeCun, S.~Denker, and S.~Solla, ``{Optimal Brain Damage},'' in {\em
  Proceedings of the International Conference on Neural Information Processing
  Systems (NIPS)}, November 1990.

\bibitem{hassibi:1993:brain_surgeon}
B.~Hassibi and D.~Stork, ``{Second Order Derivatives for Network Pruning:
  Optimal Brain Surgeon},'' in {\em Proceedings of the International Conference
  on Neural Information Processing Systems (NIPS)}, November 1993.

\bibitem{song:2015:pruning}
S.~Han, J.~Pool, J.~Tran, and W.~Dally, ``{Learning Both Weights and
  Connections for Efficient Neural Networks},'' in {\em Proceedings of the
  International Conference on Neural Information Processing Systems (NIPS)},
  December 2015.

\bibitem{song:2015:compression}
S.~Han, H.~Mao, and W.~Dally, ``{Deep Compression: Compressing Deep Neural
  Networks with Pruning, Trained Quantization and Huffman Coding},'' in {\em
  {Proceedings of the International Conference on Learning Representations
  (ICLR)}}, May 2016.

\bibitem{diannao}
T.~Chen, Z.~Du, N.~Sun, J.~Wang, C.~Wu, Y.~Chen, and O.~Temam, ``{DianNao: A
  Small-footprint High-throughput Accelerator for Ubiquitous
  Machine-learning},'' in {\em Proceedings of the International Conference on
  Architectural Support for Programming Languages and Operation Systems
  (ASPLOS)}, March 2014.

\bibitem{dadiannao}
Y.~Chen, T.~Luo, S.~Liu, S.~Zhang, L.~He, J.~Wang, L.~Li, T.~Chen, Z.~Xu,
  N.~Sun, and O.~Temam, ``{DaDianNao: A Machine-Learning Supercomputer},'' in
  {\em Proceedings of the International Symposium on Microarchitecture
  (MICRO)}, December 2014.

\bibitem{eyeriss}
Y.~Chen, T.~Krishna, J.~Emer, and V.~Sze, ``{Eyeriss: An Energy-Efficient
  Reconfigurable Accelerator for Deep Convolutional Neural Networks},'' in {\em
  Proceedings of the International Solid State Circuits Conference (ISSCC)},
  February 2016.

\bibitem{song:2015:eie}
S.~Han, X.~Liu, H.~Mao, J.~Pu, A.~Pedram, M.~Horowitz, and W.~Dally, ``{EIE:
  Efficient Inference Engine on Compressed Deep Neural Network},'' in {\em
  Proceedings of the International Symposium on Computer Architecture (ISCA)},
  June 2016.

\bibitem{eyeriss_isca}
Y.~Chen, J.~Emer, and V.~Sze, ``{Eyeriss: A Spatial Architecture for
  Energy-Efficient Dataflow for Convolutional Neural Networks},'' in {\em
  Proceedings of the International Symposium on Computer Architecture (ISCA)},
  June 2016.

\bibitem{redeye}
R.~LiKamWa, Y.~Hou, M.~Polansky, Y.~Gao, and L.~Zhong, ``{RedEye: Analog
  ConvNet Image Sensor Architecture for Continuous Mobile Vision},'' in {\em
  Proceedings of the International Symposium on Computer Architecture (ISCA)},
  June 2016.

\bibitem{minerva}
B.~Reagen, P.~Whatmough, R.~Adolf, S.~Rama, H.~Lee, S.~Lee, J.~Miguel,
  H.~Lobato, G.~Wei, and D.~Brooks, ``{Minerva: Enabling Low-Power,
  High-Accuracy Deep Neural Network Accelerators},'' in {\em Proceedings of the
  International Symposium on Computer Architecture (ISCA)}, June 2016.

\bibitem{dnn_pim_reram}
P.~Chi, S.~Li, C.~Xu, T.~Zhang, J.~Zhao, Y.~Liu, Y.~Wang, and Y.~Xie, ``{A
  Novel Processing-in-memory Architecture for Neural Network Computation in
  ReRAM-based Main Memory},'' in {\em Proceedings of the International
  Symposium on Computer Architecture (ISCA)}, June 2016.

\bibitem{isacc}
A.~Shafiee, A.~Nag, N.~Muralimanohar, R.~Balasubramonian, J.~P. Strachan,
  M.~Hu, R.~S. Williams, and V.~Srikumar, ``{ISAAC: A Convolutional Neural
  Network Accelerator with In-Situ Analog Arithmetic in Crossbars},'' in {\em
  Proceedings of the International Symposium on Computer Architecture (ISCA)},
  June 2016.

\bibitem{shidiannao}
Z.~Du, R.~Fasthuber, T.~Chen, P.~Ienne, L.~Li, T.~Luo, X.~Feng, Y.~Chen, and
  O.~Temam, ``{ShiDianNao: Shifting Vision Processing Closer to the Sensor},''
  in {\em Proceedings of the International Symposium on Computer Architecture
  (ISCA)}, June 2015.

\bibitem{nvlink}
NVIDIA, ``{NVIDIA NVLINK High-Speed Interconnect}.''
  \url{http://www.nvidia.com/object/nvlink.html}, 2016.

\bibitem{ibm_power}
IBM, ``{IBM Power Systems}.'' \url{http://www-03.ibm.com/systems/power/}, 2016.

\bibitem{alex_weird_trick}
A.~Krizhevsky, ``{One Weird Trick For Parallelizing Convolutional Neural
  Networks}.'' \url{https://arxiv.org/abs/1404.5997}, 2014.

\bibitem{dgx_1}
NVIDIA, ``{The NVIDIA DGX-1 Deep Learning System}.''
  \url{http://www.nvidia.com/object/deep-learning-system.html}, 2016.

\end{thebibliography}




\end{document}